\documentclass[lettersize,journal]{IEEEtran}
\usepackage{amsmath,amsfonts}
\usepackage{amsthm}
\usepackage{amssymb}
\usepackage{algorithmic}
\usepackage{algorithm}

\usepackage{array}
\usepackage[caption=false,font=normalsize,labelfont=sf,textfont=sf]{subfig}
\usepackage{textcomp}
\usepackage{stfloats}
\usepackage{url}
\usepackage{verbatim}
\usepackage{graphicx}
\usepackage{cite}
\usepackage{xcolor}
\usepackage{bbm}
\usepackage{makecell} 
\usepackage{fixltx2e} 

\newtheorem{definition}{Definition}

\newtheorem{lemma}{Lemma}
\newtheorem{theorem}{Theorem}
\newtheorem{corollary}{Corollary}

\newcommand{\jhedit}[1]{\textcolor{black}{#1F}}

\newcommand{\tyedit}[1]{\textcolor{black}{#1}}
\newcommand{\tyeditnew}[1]{\textcolor{black}{#1}}
\newcommand{\tyyedit}[1]{\textcolor{black}{#1}}

\newcommand{\fn}[1]{\footnote{\textcolor{black}{#1}}}

\begin{document}
\title{The Nah Bandit: Modeling User Non-compliance in Recommendation Systems}

\author{Tianyue Zhou, Jung-Hoon Cho, Cathy Wu
\thanks{Tianyue Zhou is with the Department of Civil and Environmental Engineering and the Laboratory for Information \& Decision Systems, Massachusetts Institute of Technology, Cambridge, MA 02139, USA. (e-mail: tianyuez@mit.edu)}
\thanks{Jung-Hoon Cho is with the Department of Civil and Environmental Engineering and the Laboratory for Information \& Decision Systems, Massachusetts Institute of Technology, Cambridge, MA 02139, USA. (e-mail: jhooncho@mit.edu)}
\thanks{Cathy Wu is with the Laboratory for Information \& Decision Systems; the Institute for Data, Systems, and Society; and the Department of Civil and Environmental Engineering, Massachusetts Institute of Technology, Cambridge, MA 02139, USA. (e-mail: cathywu@mit.edu)}
}

\markboth{Journal of \LaTeX\ Class Files,~Vol.~14, No.~8, August~2021}%
{Shell \MakeLowercase{\textit{et al.}}: A Sample Article Using IEEEtran.cls for IEEE Journals}

\maketitle
\begin{abstract}
Recommendation systems now pervade the digital world, ranging from advertising to entertainment. However, it remains challenging to implement effective recommendation systems in the physical world, such as in mobility or health. This work focuses on a key challenge: in the physical world, it is often easy for the user to opt out of taking \textit{any} recommendations if they are not to her liking, and to fall back to her baseline behavior. It is thus crucial in cyber-physical recommendation systems to operate with an interaction model that is aware of such user behavior, lest the user abandon the recommendations altogether.
This paper thus introduces Nah Bandit, a tongue-in-cheek reference to describe a Bandit problem where users can say `nah' to the recommendation and opt for their preferred option instead. As such, this problem lies in between a typical bandit setup and supervised learning.
We model the user non-compliance by parameterizing an anchoring effect of recommendations on users. We then propose the Expert with Clustering (EWC) algorithm, a hierarchical approach that incorporates feedback from both recommended and non-recommended options to accelerate user preference learning.
In a recommendation scenario with \(N\) users, \(T\) rounds per user, and \(K\) clusters, EWC achieves a regret bound of \(O(N\sqrt{T\log K} + NT)\), achieving superior theoretical performance in the short term compared to LinUCB algorithm.
Moreover, we show that this bound decreases further as the user compliance rate increases.
Experimental results also highlight that EWC outperforms both supervised learning and traditional contextual bandit approaches.
This advancement reveals that effective use of non-compliance feedback can accelerate preference learning and improve recommendation accuracy.
This work lays the foundation for future research in the Nah Bandit, providing a robust framework for more effective recommendation systems.
\end{abstract}

\vspace{-0.05in}
\section{Introduction}
\IEEEPARstart{O}{nline} 
recommendation systems have been widely applied in the digital world, such as personalized news recommendations \cite{li_contextual-bandit_2010}, advertisement placements \cite{pase2022rateconstrainedremotecontextualbandits}, and search engines \cite{gigli_2022_search_engine}. 
\tyedit{In such cases, a notable characteristic is that users can only access the items presented by the system.
For example, in digital shopping platforms, merchants recommend products to customers through webpages, and customers can only select from the displayed items—typically shown in a ranked list—or opt not to choose anything at all. A common approach is to model recommendation as a bandit problem, where the system selects items (arms) to recommend and learns from user feedback on recommended items to improve future suggestions \cite{bubeck2012regret, slivkins2019introduction}.} 

\tyedit{However,} they often overlook a key scenario in the physical world: users can easily opt out of taking any recommended option if it is not to their liking and revert to their baseline behavior. 
\tyedit{For instance, consider a customer shopping in a physical store—where all items are openly displayed in the showcases.
A store clerk might recommend certain items to a customer, but the customer does not always purchase the recommended ones. The customer can choose any item in the showcases that he prefers, and the clerk can observe which items the customer eventually buys. 
Examples like these are prevalent, such as in shopping \cite{kim2009personalized, wang2014modeling} and mobility recommendations \cite{jin2021endtoend,massicot2022competitive}.
It is therefore crucial for cyber-physical recommendation systems to adopt an interaction model that accounts for such behavior, leveraging non-compliance\fn{The term “non-compliance” is used solely in a descriptive sense to denote deviations from the expected or predicted behavior, and it is not intended to imply any judgment on user actions.} feedback to improve recommendations and reduce the risk of users abandoning the system entirely.}
We name this problem Nah Bandit, a tongue-in-cheek reference to describe when users say ‘nah’ to the recommendation and opt for their preferred option instead.  
\tyedit{
The Nah Bandit involves $N$ users, each interacting with the system over $T$ rounds. In each round, the system observes a user’s context (e.g., profile), selects an item to recommend, and observes the user’s actual choice from the full set of available items. Crucially, users' 
choices may be influenced by the recommendation due to the anchoring effect.}

Both supervised learning and traditional contextual bandit methods fail to address the Nah Bandit problem effectively. Supervised learning methods, such as classification based methods, decision-tree based methods, and neural networks, assume that the user selects from all options while not accounting for the influence of recommendations on users, which is called the anchoring effect.
Conversely, contextual bandit methods, such as LinUCB \cite{li_contextual-bandit_2010}, Thompson Sampling \cite{agrawal2013thompson}, and NeuralUCB \cite{zhou2020neuralcontextualbanditsucbbased}, do not incorporate feedback from non-recommended items because they assume that the user only selects from recommended options, which hinders their ability to quickly capture user preferences.
The Nah Bandit problem requires both efficiently understanding the anchoring effect and rapidly identifying user preferences from non-compliance.

\begin{table}[h!]
\caption{Comparison of Nah Bandit, traditional bandit, and supervised learning}
\label{tab:comparison2}
\centering
\begin{tabular}{|c|c|c|}
\hline
& \thead{User selects from  
\\ \textbf{recommended} options}& \thead{User selects from \\ \textbf{all} options} \\ \hline
\makecell{User is influenced \\ by recommendations}& Bandit & \makecell{Nah Bandit \\ (This work)}\\ \hline
\makecell{User is \textbf{not} influenced \\ by recommendations}& N/A & \makecell{Supervised learning}\\ \hline
\end{tabular}
\end{table}

\tyedit{The anchoring effect is a psychological phenomenon in which an individual's judgments are influenced by a reference point. In recommendation systems, it is suggested that consumers are often unconsciously misled by the information provided with the recommendations \cite{tversky1974judgment}, and thus they tend to prefer the recommended items.}
\cite{adomavicius2013recommender} found that, in the aggregate, the anchoring effect is linear and proportional to the size of the recommendation perturbation. \tyedit{Based on this,}
we propose a user non-compliance model to parameterize the anchoring effect for each user. This is the simplest method to solve the Nah Bandit problem, which reduces the bias introduced by the anchoring effect when learning user preferences. We further prove the sample complexity of parameter estimation in the user non-compliance model by transforming it into logistic regression. This result shows the speed at which we can learn user preference parameters in the Nah Bandit problem.

To rapidly capture user preferences, algorithms such as user-based Collaborative Filtering \cite{user_basedCF} use the similarity between users to make recommendations. Some approaches further assume a network structure \cite{cheng2024dynamicMAMAB, ospina2023timevarying}, 
\tyedit{or a hierarchical structure among users \cite{MALESZKA20131, ZHENG20132127} to make recommendation.}
Similar to these works, we assume a hierarchical structure in the Nah Bandit problem\tyedit{, where users' profiles exhibit a hierarchy, and users with similar profiles tend to share similar preferences}. We propose a novel hierarchical contextual bandit algorithm—Expert with Clustering (EWC).
\tyedit{EWC leverages the user non-compliance model and clustering to group users into $K$ clusters based on preference similarity.}
It then views each cluster as an expert\tyedit{, where each expert estimates the user preference as the cluster centroid. For each new user, EWC} uses the Hedge \cite{Hedge} algorithm to select the expert that best predicts user choices. The likelihood that at least one expert accurately predicts the user choice is high, regardless of compliance. 
\tyedit{This leads to rapid identification of user cluster identity.}
We further establish the regret bound of EWC. In a recommendation scenario with $N$ users, $T$ rounds per user, and $K$ user clusters, we demonstrate that EWC achieves a regret bound of $O(N\sqrt{T\log K}+NT)$. This regret bound underscores the theoretical efficacy of EWC in the short term compared to LinUCB \cite{li_contextual-bandit_2010}. 
\tyedit{Moreover, we show that this bound decreases further as the user compliance rate increases.}
We validate EWC in two different applications: travel routes and restaurant recommendations. Experimental results highlight that EWC achieves superior performance compared to both supervised learning and traditional contextual bandit approaches.

\tyedit{This paper extends and subsumes our preliminary work \cite{zhou_expert_2024}, where we introduced EWC algorithm for a travel route recommendation problem and proved a regret bound for EWC. }
However, our previous work focused on a specific travel problem that included only two options per decision, and did not formally define the Nah Bandit problem. The support vector machine (SVM) framework used for offline training in that work is not adaptable to scenarios with multiple options and, more importantly, cannot address the anchoring effect in the Nah Bandit problem. 
\tyedit{In this work, we formally introduce the Nah Bandit framework for modeling the online preference learning problem, which offers the potential to accelerate preference learning.}
We propose the user non-compliance model as the simplest method to solve the Nah Bandit problem. Compared to SVM, this model adapts to scenarios with multiple options by computing the utility of each option, and reduces the bias from the anchoring effect by parameterizing the user's dependence on recommendation.
We further combine the user non-compliance model with EWC, allowing EWC to efficiently utilize non-compliance feedback, thereby enabling rapid and accurate learning in the Nah Bandit problem.
Additionally, by incorporating user context into the preference learning process, EWC improves the speed of adapting to user preferences. 
We validate our proposed method EWC against a comprehensive set of baselines with multiple applications.
We also conduct an ablation study to assess the impact of each component. 
These enhancements and extensive evaluations help us better understand EWC algorithm and its potential applications.
Overall, this research demonstrates that effectively utilizing non-compliance feedback can accelerate preference learning and enhance recommendation accuracy. Our work establishes a foundation for future research into the Nah Bandit problem, providing a robust framework for developing more effective recommendation systems.

\begin{figure*}[!t]%
    \centering
    \includegraphics[width=0.95\textwidth]{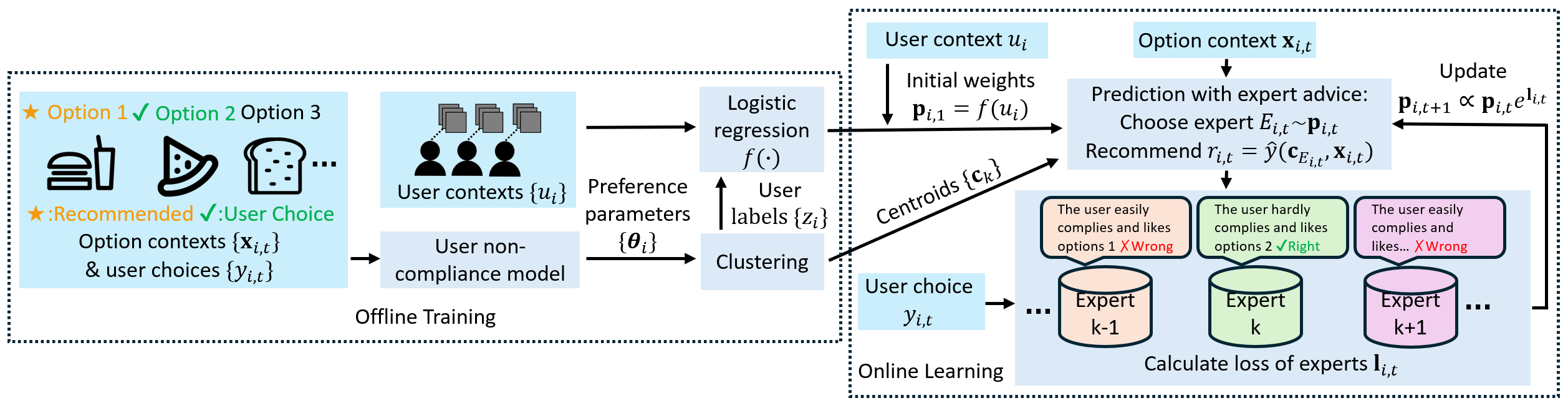}
    \caption{An overview of the Expert with Clustering (EWC) algorithm for the Nah Bandit problem. In the offline training phase, a user non-compliance model learns user preference parameters based on option contexts and user choices. These preference parameters are then grouped into clusters, with the cluster centroids serving as experts. User contexts and their cluster labels are used to train a logistic regression model to predict the initial weights of the experts. In the online learning phase, EWC selects an expert for each recommendation. After observing the user's choice, EWC calculates the losses of each experts and updates their weights accordingly.}
    \label{fig:overview_figure}
\end{figure*}
\vspace{-0.1in}
\subsection{Contributions}
We summarize our contributions as follows:
\begin{enumerate}
\item We introduce a novel bandit framework—Nah Bandit—for modeling the online preference learning problem, in which users \jhedit{can} go `nah' to the recommendation and choose their originally preferred option instead.
This framework incorporates user-observable non-compliance, offering the potential to accelerate preference learning.
\item We propose a user non-compliance model as the simplest way to solve the Nah Bandit problem, which parameterizes the anchoring effect. 
We analyze its sample complexity to show the speed at which we can learn user preference parameters.
Based on this model, we propose a hierarchical contextual bandit framework, Expert with Clustering (EWC). This framework effectively utilizes non-compliance feedback and hierarchical information, enabling rapid and accurate learning of user preferences.
\item We establish the regret bound of EWC. In a recommendation scenario with $N$ users, $T$ rounds per user, and $K$ user clusters, EWC achieves a regret bound of $O(N\sqrt{T\log K}+NT)$, achieving superior theoretical performance in the short term compared to LinUCB. \tyedit{We show that this bound decreases further as the user compliance rate increases.}
\end{enumerate}
\vspace{-0.05in}
\section{Related Works}
\subsection{Supervised Online Recommendation}
The field of recommendation systems has been significantly shaped by various supervised learning methods\tyedit{, as extensively reviewed in \cite{jannach2010recommender, aggarwal2016recommender}}. A foundational approach is Collaborative Filtering (CF), which bases its recommendations on the similarity between users \cite{user_basedCF} or items \cite{CF}. 
Matrix Factorization (MF) \cite{MF}, a specialized form of collaborative filtering, is notable for decomposing the user-item interaction matrix into lower-dimensional matrices representing the latent user and item factors.
Recent advancements have led to the introduction of online versions of these traditional methods, like Online Collaborative Filtering \cite{OnlineCF} and Online Matrix Factorization \cite{OnlineMF}, to address the dynamic nature of user preferences and real-time data.

Decision tree-based methods, such as Gradient Boosting Machines \cite{friedman2001greedy} and Random Forests \cite{breiman2001random}, and their online versions \cite{lakshminarayanan2014mondrian, beygelzimer2015online} represent another class of supervised learning methods in recommendation systems. These methods are widely used in the area of advertisements due to their ability to learn nonlinear features. 
\tyedit{XGBoost \cite{chen2016xgboost}, a fast variant of Gradient Boosting Machines with regularization techniques, is well-suited for recommendation tasks due to its ability to handle sparse data and model complex feature interactions.}

\tyedit{Same as the Nah Bandit, these supervised learning methods allows users to choose any option. However, they often assume the user's choice and feedback are independent of the recommendation, and thus overlook the potential impact of the anchoring effect.
Our research aims to address this oversight by reducing the bias introduced by the anchoring effect to make more accurate recommendations.}
\vspace{-0.1in}
\subsection{Contextual Bandits with Clustering}
The contextual bandit framework offers an efficient solution for recommendation systems. This concept was initially explored by \cite{auer_firstContxtualBandits_2002}, focusing on the use of confidence bounds within the exploration-exploitation trade-off, particularly in contextual bandits with linear value functions. 
Building on this, \cite{li_contextual-bandit_2010} extended this framework to personalized news recommendations by introducing LinUCB algorithm. \tyedit{LinUCB is simple, interpretable, and computationally efficient, establishing benchmark in this field.}
More recently, advanced algorithms such as NeuralUCB \cite{zhou2020neuralcontextualbanditsucbbased} have been developed, leveraging neural networks to model complex, non-linear relationships in the data, further enhancing the flexibility and effectiveness of contextual bandits. 
The integration of clustering techniques into the contextual bandit framework represents a significant advancement in the field. The foundational unsupervised learning algorithm, K-Means clustering, introduced by \cite{lloyd_kmeans_1982}, laid the groundwork for this development. \tyeditnew{Based on this, \cite{shepitsen2008personalized} proposed a personalized recommendation algorithm using hierarchical tag clustering. With the development of contextual bandit,} \cite{Gentile_clusterBandits_2014} proposed a novel approach to cluster users within the contextual bandit framework, using graphs to depict user similarities.
A notable progression in this area is the DYNUCB algorithm \cite{nguyen2014dynamic}, which is a dynamic clustering bandit algorithm that divides the population of users into multiple clusters and customizes the bandits for each cluster. \tyedit{This allows users to benefit from shared learning within their cluster while preserving personalization. DYNUCB also supports dynamic reassignment as user preferences evolve, enhancing adaptability.}
Building on these works, we also adopted clustering to segment users into several groups and tailor recommendations to each group.

\tyedit{However, existing contextual bandits and their clustering-based extensions cannot efficiently address the Nah Bandit problem, as they assume users choose only from the recommended options and focus solely on compliance feedback.
This has both advantages and limitations. The advantage is that, compared to supervised learning, these bandit methods implicitly capture the influence of recommendations on user behavior (i.e., the anchoring effect) as part of the reward, since the recommended options are presented to users in a consistent manner across different rounds. The limitation, however, is that they cannot leverage non-compliance feedback, and the options are not exposed to the same recommendation influence, resulting in biased user feedback.}
\tyedit{Table \ref{tab:comparison2} compares the Nah Bandit framework with traditional bandit and supervised learning approaches. The Nah Bandit offers a promising direction for accelerating preference learning by effectively incorporating non-compliance feedback. Our approach addresses this by explicitly modeling the anchoring effect and accounting for user non-compliance behavior.}
\vspace{-0.1in}

\subsection{Recommendation systems with user non-compliance} 
We refer to user non-compliance as any action taken by the user that is not among the recommended action(s).
Contextual bandit methods have been extended to consider \textit{user abandonment}, for example, deleting an app or leaving before re-engaging later.
Users may abandon the system for a variety of reasons, including \textit{fatigue} from overexposure to irrelevant content or boredom from seeing too many similar recommendations~\cite{cao2020fatigueaware}, lack of \textit{trust}~\cite{yeomans2019making}, or non-alignment with the user's immediate self-interests~\cite{bahar2020fiduciary}.
Some solutions proposed include incorporating the risk of user abandonment into the recommendation algorithms~\cite{cao2019doubly, yang2022exploration, wang2023optimizing}, or never offering recommendations that would yield lower expected reward than a user would have achieved if she acted alone~\cite{bahar2020fiduciary}.
Our work considers a softer form of user non-compliance, in which the user still selects an option within the same action class (e.g., mobility trip option), albeit not a recommended one.
Our algorithm seeks to quickly learn user preferences by acknowledging such non-compliance and learning from these user actions.
This novel approach provides a holistic view of user preferences, which is crucial for understanding the comparative utility of options and accelerating preference learning, especially in scenarios with limited data.
\vspace{-0.05in}
\section{Nah Bandit}
\subsection{Definition}
In the Nah Bandit problem, users may say `nah' to the recommendation and choose their originally preferred option instead. This means the losses (or rewards) come from user's choice. We define the Nah Bandit in a recommendation problem as follows. This framework combines elements of both supervised learning and partial feedback in a bandit setting.
\begin{definition}
    \textbf{Nah Bandit} is a scenario where a provider is tasked with recommending a set of options $O$ to a user, where $A=|O|$ represents the number of options. At each decision round $t$ within the total rounds $T$, the provider recommends one option, labeled $r_{t}\in[A]$ to the user. Subsequently, the provider observes the user's choice $y_{t}$, and incurs a loss $l_t$ from the user's choice. The objective is to minimize the cumulative regret over all decision rounds $\min_{r_{t}, \forall t}\sum_{t=1}^T l_t$.
\end{definition}
Conversely, in the traditional bandit framework for recommendation systems, the user is assumed to select only from the recommended options, and therefore, the losses (or rewards) are derived solely from those recommended options. 
\vspace{-0.1in}
\subsection{Problem Formulation}
We further incorporate contexts of users and options in the Nah Bandit. Additionally, we extend the recommendation from single-user to a multi-user. \tyyedit{We also assume a hierarchical structure among users, which is an instance of a hierarchical Bayesian model \cite{gelman1995bayesian}.} We formulate our problem as follows.

Consider a scenario where a provider is tasked with recommending a set of options, denoted as $\mathcal{O}$, to a population of users $\mathcal{U}$ with hierarchical structure, with the total number of users being $N :=|\mathcal{U}|$. Each user, identified as $i$ in the set $[N]$, is represented by a unique user context vector $u_i\in \mathbb{R}^D$. At each decision-making round $t$ within the total rounds $T$, the provider is presented with the user context $u_i$ and a specific subset of \tyedit{available} options $O_{i,t}\subset \mathcal{O}$. 
\tyedit{For simplicity, we assume that the number of available options remains constant for all subsets $O_{i,t}$, and denote this number as $A:= |O_{i,t}|$.}
Each option indexed by $a\in[A]$, is defined by an option context vector $x_{i, t, a}\in\mathbb{R}^d$. Upon receiving this information, the provider recommends one option, labeled $r_{i,t}$, from the set $[A]$ to the user. Subsequently, the provider observes the user's choice, denoted as $y_{i,t}$, and incurs a loss $l(r_{i,t}, y_{i,t})$, determined by a predefined loss function known to the provider. It is important to note that the user’s choice $y_{i,t}$ may be influenced by the recommended option $r_{i,t}$.

\tyyedit{The hierarchical structure assumes that each user's group identity $z_i$ is sampled from a group set $Z$, i.e., $z_i \sim Z$. Each user has a fixed but unknown preference parameter $\boldsymbol{\theta}_i \in \mathbb{R}^d$ that governs their decisions, where $\boldsymbol{\theta}_i \mid z_i \sim P(\cdot \mid z_i)$.}
The objective of this scenario is to minimize the total cumulative regret over all users and decision rounds. This is mathematically formulated as $\min_{r_{i,t}, \forall i,t}\sum_{i=1}^N \sum_{t=1}^T l(r_{i,t}, y_{i,t})$.
\vspace{-0.05in}
\section{A User Non-compliance Model}
\subsection{Model Description}
A key assumption in our problem is that the user's choice may be influenced by the anchoring effect. This leads to a scenario of partial feedback akin to a contextual bandit setting, where learning user preferences can be challenging. \cite{adomavicius2013recommender} uses a rating drift, defined as the actual rating minus the predicted rating, to represent the anchoring effect. They found that, in aggregate, the rating drift is linear and proportional to the size of the recommendation perturbation. This means that the more we recommend one option, the higher rating the user will give to this option. However, the slope of this linear relationship, which represents the user's additional preference for the recommended options, is unknown. 
Building on \cite{adomavicius2013recommender}, we propose a user non-compliance model to discern this additional preference
\tyedit{and thereby address the anchoring effect.}

\tyyedit{Assume that, given preference parameter $\boldsymbol{\theta}_{i}$, we can make predictions using a known function $\hat{y}(\boldsymbol{\theta}_{i}, \mathbf{x}_{i,t})$.} The key idea is that we assume there exists a $\theta_{i,\text{rec}}$ within $\boldsymbol{\theta}_{i}$ that quantifies the additional preference toward recommended options. Users with a high $\theta_{i,\text{rec}}$ highly rely on the recommended option, while users with $\theta_{i,\text{rec}}=0$ select the option with the highest utility for them, regardless of the recommendation. Our goal is to learn this $\theta_{i,\text{rec}}$ for each user.

We propose a user non-compliance model, which is a linear model that parameterizes the user's dependence on the recommendation.
First, we incorporate a context $x^{\text{rec}}_{i,t,a}\in \mathbb{R}$ in each option context $x_{i,t,a}$ which represents the degree to which this item is recommended to the user. For example, $x^{\text{rec}}_{i,t,a}=\mathbbm{1}_{r_{i,t}=a}$.
The utility of each option is then defined as $U_{i,t,a}=x_{i,t,a}^{\intercal} \boldsymbol{\theta}_i$. Let $\mathbf{U}_{i,t}=[U_{i,t,1}, U_{i,t,2}, \ldots, U_{i,t,A}]$ represent the utility vector. The probability of selecting each option is predicted as $\mathbf{p}_{i,t}=\boldsymbol{\sigma}(\mathbf{U}_{i,t})$, where $\boldsymbol{\sigma}(\cdot)$ denotes the softmax function. Let $\mathbf{y}_{i,t}$ be the one-hot encoding of $y_{i,t}$.
The discrepancy between the predicted probability and the actual choice is quantified using the KL-divergence. The detailed methodology is encapsulated in Algorithm \ref{Supervised_Preference_Learning}.

This approach is the simplest way to solve the Nah Bandit problem. It provides a supervised learning way to learn user's preference parameters $\theta_i$. It reduces the bias in the learning process that might be introduced by the anchoring effect, thereby preventing the user non-compliance model from falling into sub-optimal recommendations. 
\tyedit{In Section \ref{sec:EWC}, we further introduce how the user non-compliance model is integrated into our main approach to accelerate preference learning in the Nah Bandit problem.}

\begin{algorithm}
    \caption{User Non-compliance Model}
    \label{Supervised_Preference_Learning}
    \begin{algorithmic}
        \REQUIRE Option contexts $\{x_{i,t,a}\}_{i\in[N],t\in[T],a\in[A]}$, recommendation record $\{r_{i,t}\}_{i\in[N], t\in[T]}$, user choice $\{y_{i,t}\}_{i\in[N],t\in[T]}$
        \STATE $x_{i,t,a} \leftarrow[x_{i,t,a},x^{\text{rec}}_{i,t,a}]$ for all $i\in[N],t\in[T],a\in[A]$
        \STATE Randomly initialize $\{\boldsymbol{\theta}_i\}_{\forall i \in [N]}$
        \WHILE{$\{\boldsymbol{\theta}_i\}_{i \in [N]}$ not converge}
            \STATE $U_{i,t,a}\leftarrow x_{i,t,a}^{\intercal} \boldsymbol{\theta}_i$ for all $i\in[N],t\in[T],a\in[A]$
            \STATE $\mathbf{p}_{i,t}\leftarrow \boldsymbol{\sigma}(\mathbf{U}_{i,t})$ for all $i\in[N],t\in[T]$
            \STATE $\boldsymbol{\theta}_{i}\leftarrow\arg\min_{\boldsymbol{\theta}} \frac{1}{T}\sum_{t=1}^T \text{KL}(\mathbf{p}_{i,t}||\mathbf{y}_{i,t})$ for all $i\in[N]$
        \ENDWHILE
    \end{algorithmic}
\end{algorithm}
\vspace{-0.1in}
\subsection{Sample Complexity Analysis}
This section presents a sample complexity analysis of user preference parameter estimation in the user non-compliance model. 
If we let the model use the context of only two options to update the model, where one is the user's choice and the other is not, this model is equivalent to a logistic regression. 
\tyedit{To facilitate analysis, in Theorem \ref{sample_non-compliance} we focus on the two-option update setting and derive the sample complexity of the user non-compliance model using known results from logistic regression. While this approach does not make use of all available feedback, it is practical in experimental settings and guarantees the theoretical bound. In real-world applications, leveraging feedback from all available options typically results in better performance.}
\begin{lemma}[Sample Complexity of Parameter Estimation in Logistic Regression (Theorem 4 in \cite{hsu2024samplecomplexityparameterestimation})]\label{sample_LR} 
    Consider a logistic regression model with input $x\in\mathbb{R}^d\sim \mathcal{N}(\mathbf{0}, \mathbf{I}_d)$ and output $y\in\{-1,1\}$. The parameter space is the unit sphere $S^{d-1}=\{\theta\in\mathbb{R}^d: ||\theta||=1\}$. $y|x\sim\text{Bern}(\sigma(\beta x^\intercal \theta^*))$ where $\sigma$ is the Sigmoid function, $\beta$ is the inverse temperature, and $\theta\in S^{d-1}$ is the parameter of the model. The observed data $\{x_t,y_t\}_{t=1}^T$ are independent copies of $x$ and $y$ with unknown parameter $\theta^*\in S^{d-1}$. For any fixed $\epsilon,\delta\in(0,1)$, assume $\beta\geq \frac{4\sqrt{2\pi}}{\epsilon}$, and $T\geq \frac{C(d\log(1/\epsilon)+\log(1/\delta))}{\epsilon}$ where $C>0$ is an absolute constant. Then with probability at least $1-\delta$, the empirical risk minimizer $\hat{\theta}_{\text{ERM}}(\{x_t, y_t\}_{t=1}^T)$ achieves $||\hat{\theta}_{\text{ERM}}(\{x_t, y_t\}_{t=1}^T)-\theta^*||\leq \epsilon$.
\end{lemma}
\begin{theorem}[Sample Complexity of Parameter Estimation in the User Non-compliance Model]\label{sample_non-compliance}
    Let the user non-compliance model use the context of only two options to update the model, where one is the user’s choice and the other is not. Assume $\tilde{x}_{i,t}:=x_{i,t,1}-x_{i,t,2}\sim \mathcal{N}(\mathbf{0},\mathbf{I}_d)$. Suppose $\{\tilde{x}_{i,t}, y_{i,t}\}$ are i.i.d. samples from a a distribution determined by $\theta_i^*\in S^{d-1}$, where $y_{i,t}|\tilde{x}_{i,t}\sim \text{Bern}(\sigma(\beta_i \tilde{x}_{i,t}^\intercal \theta_{i}^*))$. Assume the user non-compliance model outputs the empirical risk minimizer $\theta_i=\hat{\theta}_{\text{ERM}}(\{x_{i,t}, y_{i,t}\}_{t=1}^T)$. For any fixed $\epsilon,\delta\in(0,1)$, if $\beta_i\geq \frac{4\sqrt{2\pi}}{\epsilon}$, and $T\geq \frac{C(d\log(1/\epsilon)+\log(1/\delta))}{\epsilon}$ where $C>0$ is an absolute constant, then with probability at least $1-\delta$, we have $||\theta_i-\theta^*_i||\leq \epsilon$.
\end{theorem}
Lemma \ref{sample_LR} shows the sample complexity of parameter estimation in a logistic regression model. Using Lemma \ref{sample_LR}, we can get Theorem \ref{sample_non-compliance}, which shows the sample complexity of the user preference parameter in the user non-compliance model. The proof of Theorem \ref{sample_non-compliance} is in Appendix \ref{Proof_sample}. This result demonstrates the speed at which we can learn user preference parameters in the Nah Bandit problem. 
\vspace{-0.1in}
\section{Expert with Clustering}\label{sec:EWC}
\subsection{General Framework}
Another core aspect of our problem is rapidly identifying user preferences based on both compliance and non-compliance. To address this, we introduce the Expert with Clustering (EWC) algorithm, a novel hierarchical contextual bandit approach. 
EWC consists of both an offline training phase and an online learning phase. It transforms the recommendation problem into a prediction with expert advice problem, using clustering to get experts during the offline training phase and employing the Hedge algorithm to select the most effective expert during the online learning phase.

Prediction with expert advice is a classic online learning problem introduced by \cite{Hedge}. Consider a scenario in which a decision maker has access to the advice of $K$ experts. At each decision round $t$, advice from these $K$ experts is available, and the decision maker selects an expert based on a probability distribution $\mathbf{p}_{t}$ and follows his advice. Subsequently, the decision maker observes the loss of each expert, denoted as $\mathbf{l}_{t} \in [0,1]^K$. The primary goal is to identify the best expert in hindsight, which essentially translates to minimizing the regret: $\sum_{t=1}^T\left(<\mathbf{p}_{t},\mathbf{l}_{t}>- \mathbf{l}_{t}(k^*)\right)$, where $k^*$ is the best expert throughout the time.

We cast the recommendation problem into the framework of prediction with expert advice in the following way.
Recall the assumption that each user has a fixed but unknown preference parameter $\boldsymbol{\theta}_{i} \in \mathbb{R}^d$. Given $\boldsymbol{\theta}_{i}$, EWC algorithm operates under the assumption of a cluster structure within the users' preference parameters $\{\boldsymbol{\theta}_{i}\}_{i\in[N]}$. 

In the offline training phase, utilizing a set of offline training data $\mathcal{D}= \{ \{x_{i,t, a}\}_{i\in[N^\prime], t\in[T], a\in[A]}, \{y_{i,t}\}_{i\in[N^\prime],t\in[T]} \} $ where $N^\prime$ and $T^\prime$ are number of users and decision rounds in training data, we initially employ the user non-compliance model as a learning framework to determine each user's preference parameter $\boldsymbol{\theta}_{i}$.
Despite differences between training and testing data, both are sampled from the same distribution. This allows for an approximate determination of \(\boldsymbol{\theta}_{i}\), providing insights into the hierarchical structure among users, albeit with some degree of approximation.
Subsequently, a clustering method is applied on $\{\boldsymbol{\theta}_i\}_{i\in[N^\prime]}$ to identify centroids $\{\mathbf{c}_{k}\}_{k\in[K]}$ and user's cluster affiliation $\{z_{i,k}\}_{i\in[N^\prime],k\in[K]}$, where $\mathbf{c}_k\in\mathbb{R}^d$ and $z_{i,k}\in\{0,1\}$, $\sum_{k\in[K]}z_{i,k}=1$. The number of clusters $K$ serves as a hyperparameter. 
\tyedit{To select $K$, we evaluate different values on the offline training set and choose the one that yields the minimum regret for the EWC algorithm.}

Each centroid is considered an expert. 
In \jhedit{the} online learning phase, using the Hedge algorithm, we initialize their weights and, at every online decision round, select an expert $E_{i,t} \in [K]$. An expert $E_{i,t}$ provides advice suggesting that a user's preference parameters closely resemble the centroid $\mathbf{c}_{E_{i,t}}$. Consequently, we use this centroid to estimate the user’s preferences. The recommendation $r_{i,t} = \hat{y}(\mathbf{c}_{E_{i,t}}, \mathbf{x}_{i,t})$ is then formulated. For example, $\hat{y}(\boldsymbol{\theta}, \mathbf{x}_{i,t})=\arg\max_{a} x_{i,t,a}^\intercal \boldsymbol{\theta}$ where $\mathbf{x}_{i,t}=[x_{i,t,1}, x_{i,t,2}, ...,x_{i,t,A}]$.
Upon receiving the user's chosen option $y_{i,t}$, we calculate the loss for each expert and update the weights in Hedge based on this loss.
The loss for each expert $k$ is determined by a known loss function $\mathbf{l}_{i,t}(k) = l(\hat{y}(\mathbf{c}_{k}, \mathbf{x}_{i,t}), y_{i,t}) \in \mathbb{R}$. For example, $l(\hat{y}(\mathbf{c}_{k}, \mathbf{x}_{i,t}), y_{i,t})=\mathbbm{1}_{\hat{y}(\mathbf{c}_{k}, \mathbf{x}_{i,t})\neq y_{i,t}}$.
The details of this process are encapsulated in Algorithm \ref{Expert_with_Cluster}.

The EWC algorithm efficiently utilizes non-compliance feedback in both the offline training and the online learning phase. In offline training, the learning framework within EWC is an interchangeable module that can be implemented using various models, such as SVM or neural networks. Compared to other models, the user non-compliance model captures preferences from both compliance and non-compliance feedback with low bias. 
\tyedit{In online learning, EWC leverages the user non-compliance model to transform the preference learning problem into identifying a new user's cluster identity. Given the high likelihood that at least one cluster centroid accurately predicts the user's preferences—regardless of compliance—EWC efficiently infers the user's cluster identity.} 

\begin{algorithm}
    \caption{Expert with Clustering}
    \label{Expert_with_Cluster}
    \begin{algorithmic}
        \REQUIRE Number of clusters $K$, offline training data $\mathcal{D}$, learning rate $\eta$
        \STATE Train data by Algorithm \ref{Supervised_Preference_Learning}, receive $\{\boldsymbol{\theta}_{i}\}_{i\in[N^\prime]}$
        \STATE Apply clustering on $\{\boldsymbol{\theta}_{i}\}_{i\in[N^\prime]}$, receive centroids $\{\mathbf{c}_{k}\}_{k\in [K]}$ 
        \STATE Initialize weight $\mathbf{p}_{i,1}(k) \leftarrow \frac{1}{K}$ for all $i\in[N], k\in[K]$
        \FOR {$t = 1,\dots,T$}
            \FOR {$i = 1,\dots,N$}
                \STATE Receive $\mathbf{x}_{i,t}$ 
                \STATE Sample $E_{i,t} \sim \mathbf{p}_{i,t}$, submit $r_{i,t} \leftarrow \hat{y}(\mathbf{c}_{E_{i,t}}, \mathbf{x}_{i,t})$, 
                \STATE Receive $y_{i,t}$, compute loss $\mathbf{l}_{i,t}(k) \leftarrow l(\hat{y}(\mathbf{c}_{k}, \mathbf{x}_{i,t}), y_{i,t})$ for all $k\in[K]$
                \STATE $\mathbf{p}_{i,t+1}(k) \leftarrow \frac{\mathbf{p}_{i,t}(k)e^{-\eta \mathbf{l}_{i,t}(k)}}{\sum_{k^\prime \in [K]} \mathbf{p}_{i,t}(k^\prime)e^{-\eta \mathbf{l}_{i,t}( k^\prime)}}$ for all $k\in[K]$
            \ENDFOR
        \ENDFOR
    \end{algorithmic}
\end{algorithm}
\vspace{-0.05in}
\subsection{Clustering with Loss-guided Distance}
The core parameter influencing the regret in our model is the set of centroids \(\{\mathbf{c}_{k}\}_{k\in [K]}\). An accurately representative set of centroids can significantly reflect users' behaviors, whereas poorly chosen centroids may lead to suboptimal performance. In our simulations, we observed that the standard K-Means algorithm has limitations, as similar \(\boldsymbol{\theta}_{i}\) values in the Euclidean space do not necessarily yield similar user preferences. 

To address the limitation of K-Means clustering, researchers in fields such as federated learning \cite{ICFA_NEURIPS2020, FMTL_IEEE2019} and system identification \cite{clustered_system_CDC2023} have devised bespoke objective functions to enhance clustering methodologies. Inspired by \cite{ICFA_NEURIPS2020},
we introduce a distance metric guided by the loss function which is tailored for online preference learning. Our objective is to ensure that \(\theta_i\) values within the same cluster exhibit similar performance. Thus, we replace the traditional L$_2$ norm distance with the prediction loss incurred when assigning \(\mathbf{c}_{k}\) to user \(i\). Here, we define: \(\mathbf{X}_i=[\mathbf{x}_{i,1}, \mathbf{x}_{i,2},...,\mathbf{x}_{i,T^\prime}] \in \mathbb{R}^{T^\prime\times A \times d}\), $\mathbf{\hat{y}}(\mathbf{c}_{k}, \mathbf{x}_{i,t})$ be the one-hot encodings of $\hat{y}(\mathbf{c}_{k}, \mathbf{x}_{i,t})$, \(\mathbf{y}_i=[\mathbf{y}_{i,1}, \mathbf{y}_{i,2},...,\mathbf{y}_{i,T^\prime}] \in \mathbb{R}^{T^\prime \times A}\), and \(\mathbf{\hat{y}}(\mathbf{c}_{k}, \mathbf{X}_i)= [\mathbf{\hat{y}}(\mathbf{c}_{k}, \mathbf{x}_{i,1}), \mathbf{\hat{y}}(\mathbf{c}_{k}, \mathbf{x}_{i,2}),..., \mathbf{\hat{y}}(\mathbf{c}_{k}, \mathbf{x}_{i,T^\prime})] \in \mathbb{R}^{T^\prime \times A} \). The Loss-guided Distance is defined as $\text{dist}(i, \mathbf{c}_{k})=||\mathbf{\hat{y}}(\mathbf{c}_{k}, \mathbf{X}_i)-\mathbf{y}_i||^2_F$.
The detailed clustering is presented in Algorithm~\ref{K-Means with Loss-guided Distance}. 

\begin{algorithm}
	\caption{K-Means with Loss-guided Distance}
	\label{K-Means with Loss-guided Distance}
	\begin{algorithmic}
		\REQUIRE $\{\boldsymbol{\theta}_{i}\}_{i\in [N^\prime]}$
        \STATE Randomly initialize centroids $\{\mathbf{c}_{k}\}_{k\in[K]}$
        \WHILE{$\{\mathbf{c}_{k}\}_{k\in[K]}$ not converged}
        \STATE $\text{dist}(i, \mathbf{c}_{k})\leftarrow ||\mathbf{\hat{y}}(\mathbf{c}_{k}, \mathbf{X}_i)-\mathbf{y}_i||^2_F$ for all $i\in[N^\prime], k\in[K]$
        \STATE $z_{i,k}\leftarrow \mathbbm{1}_{k=\arg\min_{k^\prime} \text{dist}(i,c_{k^\prime})}$ for all $i\in[N^\prime],k\in[K]$
        \STATE $\mathbf{c}_{k}\leftarrow \frac{\sum_{i=1}^{N}z_{i,k}\boldsymbol{\theta}_{i}}{\sum_{i=1}^N z_{i,k}}$ for all $k\in[K]$
        \ENDWHILE
        \RETURN $\{\mathbf{c}_{k}\}_{k\in[K]}$, $\{z_{i,k}\}_{i\in[N^\prime], k\i n[K]}$
	\end{algorithmic}
\end{algorithm}
\vspace{-0.05in}
\subsection{Accelerating Learning with User Context}
In our model, we capitalize on the user context to facilitate accelerated preference learning during the initial phase. We hypothesize a latent relationship between the user context and the user's cluster affiliation. In the offline training phase, we utilize user context vectors $\{u_{i}\}_{i\in[N^\prime]}$ along with the users' cluster labels $\{z_{i,t}\}_{i\in[N^\prime], t\in[T]}$ to train a logistic regression model, denoted as $f:\mathbb{R}^D\rightarrow \mathbb{R}^K$. This model is designed to map the user context to a probabilistic distribution over the potential cluster affiliations.

During the online learning phase, we employ the trained logistic regression model $f(\cdot)$ to predict the probability of each user’s group affiliation based on their respective context. These predicted probabilities are then used to initialize the weights of the experts for each user, i.e., $\mathbf{p}_{i,1}\leftarrow f(u_i)$ for all $i\in[N]$.
\vspace{-0.1in}
\section{Regret Analysis}
\subsection{Regret Bound of EWC}
In our problem, we define the loss function $l(\hat{y}(\mathbf{c}_{k}, \mathbf{x}_{i,t}), y_{i,t})=\mathbbm{1}_{\hat{y}(\mathbf{c}_{k}, \mathbf{x}_{i,t})\neq y_{i,t}}$. We define the regret of EWC as the performance difference between EWC and recommendation with known user preference parameter $\boldsymbol{\theta}_i$:
\begin{equation}
    R_{\text{EWC}} = \sum_{i=1}^N \sum_{t=1}^T \left( \langle \mathbf{p}_{i,t}, \mathbf{l}_{i,t}\rangle - 
    l(\hat{y}(\boldsymbol{\theta_i},\mathbf{x}_{i,t}), y_{i,t})
    \right)
\end{equation}
Since the study in \cite{K-MeansBound} shows the performance of K-Means clustering using the $L_2$ norm distance, we similarly adopt the $L_2$ norm distance to analyze regret in our framework.
Theorem \ref{Regret_EWC} is our main theoretical result \tyedit{which shows the regret bound of EWC algorithm}. The proof is in Appendix \ref{Proof_EWC}. 

\newtheorem{EWC}{Theorem}
\begin{theorem}[Regret Bound of EWC]\label{Regret_EWC}
Let $P$ be any distribution of $\boldsymbol{\theta}_{i}\in\mathbbm{R}^d$ with $\boldsymbol{\mu} = \mathbbm{E}_P[\boldsymbol{\theta}_{i}]$, $\sigma^2 =\mathbbm{E}_P[||\boldsymbol{\theta}_{i}-\boldsymbol{\mu}||^2]$, and finite \textit{Kurtosis}. Let 
$k^*(i)$ be the optimal expert for user $i$, and $\mathcal{L}=\sum_{i=1}^N ||\mathbf{c}_{k^*(i)}-\boldsymbol{\theta}_{i}||^2$
. If $\mathbf{\hat{y}}(\cdot, \mathbf{X}_i)$ is Lipschitz continuous for all $\mathbf{X}_i$ with Lipschitz constant $L$, Frobenius distance, and dimension normalization, then with probability at least $1-\delta$, the regret of EWC is bounded by:
\begin{equation}
    R_{\text{EWC}} \leq 2N\sqrt{T\log K}
        +\frac{1}{4}TL\left(\epsilon\sigma^2+(\epsilon+2)\mathbb{E}[\mathcal{L}]\right)
\end{equation}
\end{theorem}

\tyyedit{Here the Lipschitz condition here means that $\exists L$ s.t. $\forall i, \forall \boldsymbol{\theta}_1,\boldsymbol{\theta}_2$, $\frac{1}{T}||\mathbf{\hat{y}}(\boldsymbol{\theta}_1, \mathbf{X}_i)-\mathbf{\hat{y}}(\boldsymbol{\theta}_2,\mathbf{X}_i)||^2_F\leq L ||\boldsymbol{\theta}_1-\boldsymbol{\theta}_2||^2$. This condition implies that a user's predicted choices do not vary drastically with small changes in their preference parameter.} \tyedit{Theorem \ref{Regret_EWC} indicates that the regret bound of EWC consists of two components. 1) \( 2N\sqrt{T\log K} \), reflects the regret incurred from using the Hedge algorithm to identify a user's cluster identity. 2) \( \frac{1}{4}TL\left(\epsilon\sigma^2 + (\epsilon+2)\mathbb{E}[\mathcal{L}]\right) \), captures the bias introduced by approximating user preferences with cluster centroids. A lower clustering loss \( \mathcal{L} \) directly reduces this component, thereby lowering the overall regret.
The users’ compliance rates \( \theta_{i,\text{rec}} \) further influence the regret by affecting the Lipschitz constant \( L \), which characterizes how sensitively a user’s choices \( \mathbf{\hat{y}}(\cdot, \mathbf{X}_i) \) change in response to variations in their preference parameters \( \boldsymbol{\theta}_i \). When \( \theta_{i,\text{rec}} \) is large, the recommended option tends to dominate the user’s decision, making them more likely to choose it regardless of other parameter values. This leads to smaller differences in user behavior and, consequently, a smaller Lipschitz constant \( L \) which significantly reduces the upper bound on regret.}

The Gaussian Mixture Model (GMM) \tyedit{\cite{Hasselblad01081966}} aligns closely with our hypothesis of a hierarchical structure among users, which is a typical assumption in the analysis of clustering algorithms. By assuming that the distribution of users' preferences follows a GMM, we derive Corollary~\ref{EWC_Bound_GMM}. 
\begin{corollary}\label{EWC_Bound_GMM}
    Assume $P$ is a Gaussian Mixture Model (GMM) \tyedit{$\sum_{k=1}^K \pi_k \mathcal{N}(\mu_k, \Sigma_k)$} with $K$ Gaussian distributions, each of which has weight $\pi_k$, mean $\boldsymbol{\mu}_k$, and covariance $\Sigma_k$, and the clustering outputs the optimal centroids where $\mathbf{c}_{k}=\boldsymbol{\mu}_k$. Define 
$l_{\text{centroids}}=\frac{1}{4N}L\epsilon \sigma^2 +\frac{1}{4}L(\epsilon+2)\sum_{k=1}^K \pi_k \text{trace}(\Sigma_k)$
be the average loss caused by centroids. With probability at least $1-\delta$, the regret of EWC is bounded by
\begin{equation}
    R_{\text{EWC}}\leq \overline{R}_{\text{EWC}}= 2N\sqrt{T\log K} +  TNl_{\text{centroids}}
\end{equation}
\end{corollary}
\tyeditnew{The proof of Corollary \ref{EWC_Bound_GMM} is provided in Appendix \ref{Proof_EWC_Coro}. EWC does not achieve sublinear regret in the long term because it uses the cluster centroid as an estimate of user-specific preferences each time. However, if the estimation error is low, indicated by a small $l_{\text{centroids}}$ value, EWC achieves low regret in the short term.}
\vspace{-0.1in}
\subsection{Comparison to LinUCB}
\tyeditnew{
\tyedit{\cite{chu2011contextual} showed that, with probability at least \( 1 - \delta \), the regret bound of SupLinUCB (a variant of LinUCB) is \(\overline{R}_{\text{LinUCB}}= O\left(N\sqrt{Td\log^3{(AT\log T/\delta)}}\right) \). Building on this result, we derive Corollary \ref{AdvEWC}, which compares the regret bounds of EWC and LinUCB.}
The proof of Corollary \ref{AdvEWC} is provided in Appendix \ref{Proof_AdvEWC}. EWC demonstrates superior theoretical performance compared to LinUCB when $T$ is relatively small.}
\begin{corollary}[Advantage of EWC]\label{AdvEWC}
Assume $\overline{R}_{\text{LinUCB}}= CN\sqrt{Td\log^3{(AT\log T/\delta})}$, then when $T< (\frac{C-2}{l_{\text{centroids}}})^2$, $\overline{R}_{\text{EWC}}<\overline{R}_{\text{LinUCB}}$.
\end{corollary}
\vspace{-0.05in}
\section{Experiment}
In this section, we perform empirical analysis to validate our algorithm in two different applications: travel routes and restaurant recommendations. 
\vspace{-0.1in}
\subsection{Baselines}
We compare our EWC algorithm against several baseline methods to determine its performance in online preference learning.
The \textit{user non-compliance model (Non-compliance)} is a linear model that parameterizes user dependence on recommendation (Algorithm \ref{Supervised_Preference_Learning}).
\textit{LinUCB}
refines the upper confidence bound method to suit linear reward scenarios, aiming to strike a balance between exploring new actions and exploiting known ones. We adopt the hybrid linear model in LinUCB proposed by \cite{li_contextual-bandit_2010} to learn from both user context and option context. 
\textit{DYNUCB} combines LinUCB with dynamic clustering, which divides the population of users into multiple clusters and customizes the bandits for each cluster. We also use the hybrid linear model in DYNUCB.
\textit{XGBoost} is a highly efficient supervised learning algorithm based on gradient boosting.
\vspace{-0.05in}
\subsection{Travel Route Recommendation}
We validate our algorithm in travel route recommendations. The data is collected from a community survey first, and then expanded to represent a diverse driving population.
\subsubsection{Description}
Consider a social planner is tasked with recommending travel route options to a population of drivers, where each driver $i$ has a user context vector $u_i$. Each route option at decision round $t$ with index $a$ is parameterized by an option context vector $x_{i,t,a}$. For simplicity, we consider two options ($A=2$), each with two relevant travel metrics ($d=2$): travel time and emission. \tyedit{At each decision round of a user,} the social planner faces a choice between two options: route 1, the standard route with regular travel time and emissions, and route 2, an eco-friendly alternative that offers reduced emissions while coming with increased travel time.
\subsubsection{Experimental Setup}
\textbf{Community Survey. }
This study involved a community survey conducted in July 2023 on the University of North Carolina at Charlotte campus, and a total of 43 individuals participated.
Participants provided the driving choice preferences as well as demographic data covering age, gender, ethnicity, and educational level.
The survey's main component involved a series of questions assessing willingness to adhere to route recommendations under varying scenarios with distinct travel times and carbon dioxide emission levels. Participants rated their likelihood of following these recommendations on the Likert scale, offering insight into their decision-making criteria.
\textbf{Mobility User Simulation}.
To better represent a diverse driving population, we expanded our dataset.
We use the Bayesian inference model that resembles the original distribution from the survey data \cite{andrieu_introduction_2003}.
For each user in the survey data, we sample his preference parameter $\boldsymbol{\theta}_i$ from a multivariate normal distribution. 
Based on this $\boldsymbol{\theta}_i$, we calculate the prediction loss $\mathcal{L}$ compared to the real likelihood. Then we calculate the acceptance rate $p=e^{-\lambda \mathcal{L}}$. We accept this sample with probability $p$. The process above is repeated until we collect $24$ samples for each user. In order to incorporate the influence of the recommendation on the user's choice, we concatenate $\boldsymbol{\theta}_i$ with $\theta_{\text{rec}}$ where $\theta_{\text{rec}}$ is sampled from a beta distribution and then multiplied by $\beta$. 
Higher $\beta$ means the population has more preference for the recommended option.
$\beta=0$ represents a supervised learning scenario where the user's choice is independent of the recommendation, while $\beta>0$ means a bandit feedback scenario.
Based on the synthetic preference parameters, we sample travel routes and generate user choices on users' routes. The detailed context description and parameter setting are shown in Appendix \ref{details_exp}.

\subsubsection{Results and Interpretation}\label{Results and Interpretation}
\begin{figure*}[!t]%
    \centering
    \hfill
    \subfloat[$\beta=0$]{\includegraphics[width=0.33\textwidth]{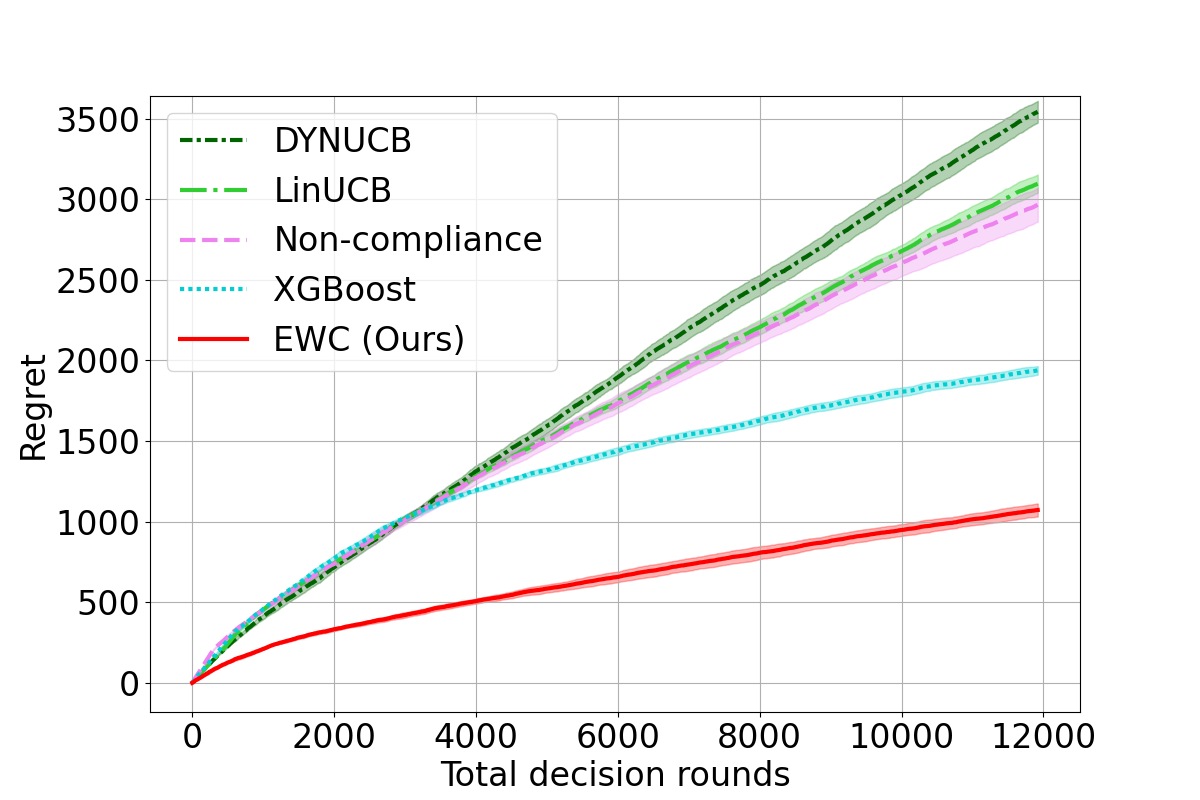}%
    \label{fig:beta=0}}
    \hfill
    \subfloat[$\beta=1$]{\includegraphics[width=0.33\textwidth]{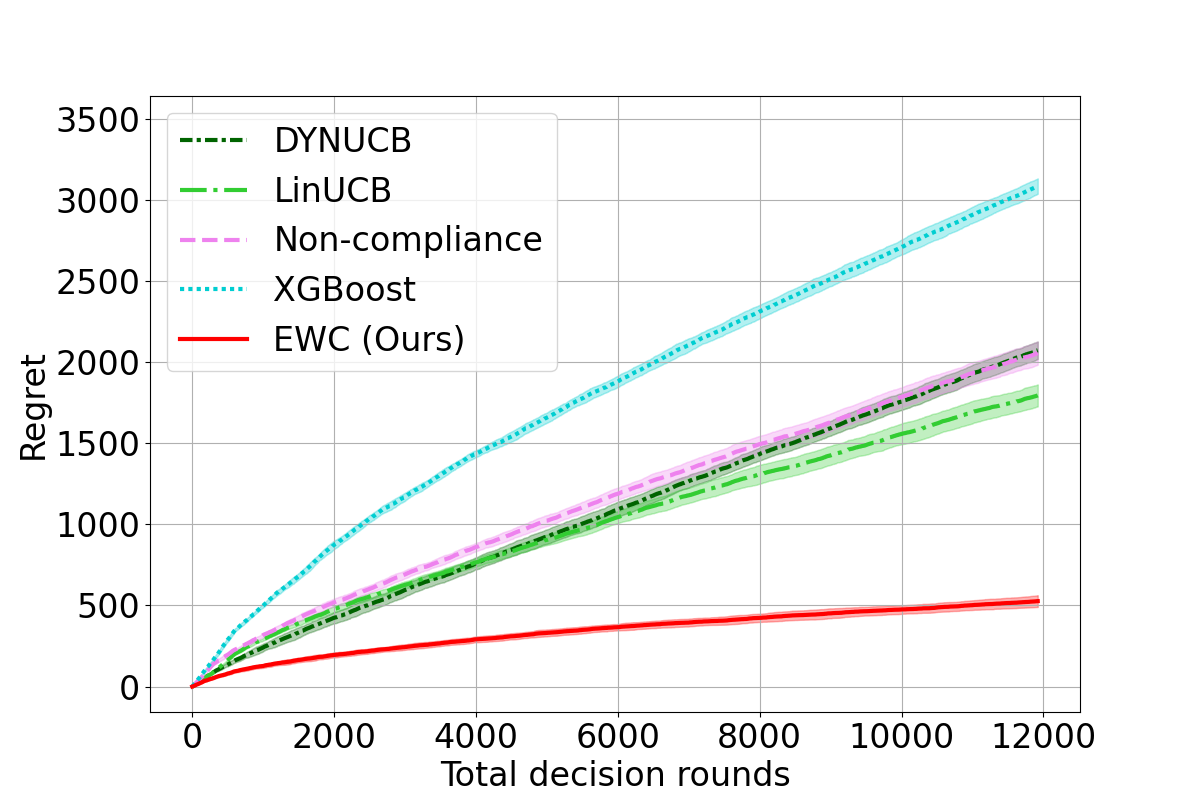}%
    \label{fig:beta=1}}
    \hfill
    \subfloat[$\beta=10$]{\includegraphics[width=0.33\textwidth]{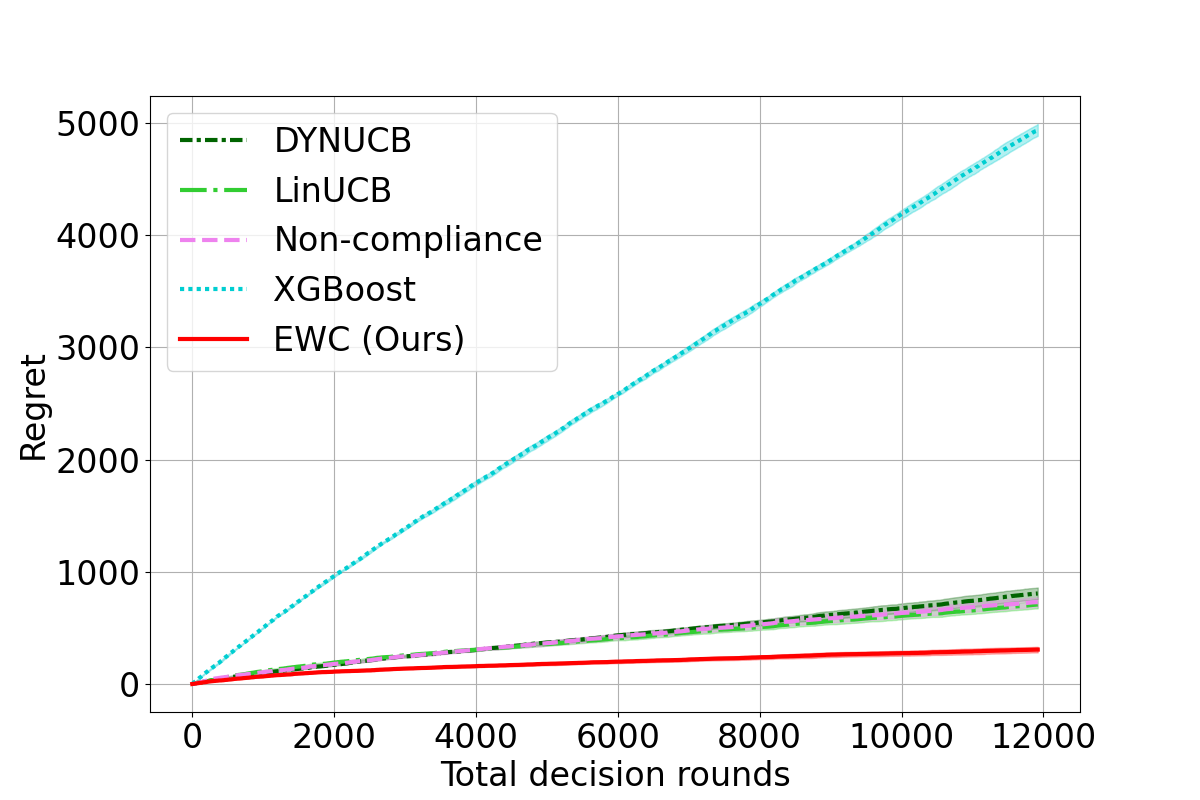}%
    \label{fig:beta=10}}
    \hfill
    \caption{Regret of Expert with Clustering (EWC, Ours) and other baselines (DYNUCB, LinUCB, the user non-compliance model, and XGBoost) applied to travel route recommendation data. The x-axis represents the total decision rounds, while the y-axis represents the regret. \tyeditnew{Lower regret indicates better performance.} The comparison includes three scenarios with different values of $\beta$, where $\beta$ indicates the scale of users' dependence on recommendations in data generation. A higher $\beta$ means the population has a stronger preference for the recommended option. EWC consistently shows significantly lower regret than other baseline methods across all scenarios.}
    \label{fig:travel}
\end{figure*}

In this section, we present the relative performance of our proposed algorithm, EWC, by comparing it with various baselines over a series of $12,000$ total rounds. The experiment is repeated with 5 different random seeds. Figure~\ref{fig:travel} shows the results with the travel route recommendation dataset. The regret represents the cumulative difference between the rewards of the algorithm's selections and the optimal choices.

Our proposed algorithm, EWC, demonstrates a significantly lower regret than that of other baseline methods in all scenarios. It achieves a very low slope in the early rounds, indicating that EWC algorithm effectively incorporates user preference information and rapidly learns user preferences.

The user non-compliance model can learn user’s preference from both compliance and non-compliance feedback, as its slope increasingly decreases. However, it does not learn rapidly, as its slope decreases slowly and it does not show a significant advantage over LinUCB. 
Compared to the user non-compliance model, EWC achieves a much lower regret. This is because EWC efficiently uses the hierarchical information within the group of users' preferences learned by the user non-compliance model. It accelerates the preference learning process by clustering and prediction with expert advice.

XGBoost achieves the second lowest regret when $\beta=0$, which represents a supervised learning scenario. However, as $\beta$ increases, representing a bandit feedback scenario where users have a stronger preference for recommended routes, XGBoost’s performance deteriorates significantly. This suggests that supervised learning methods overlook the influence of recommendations on user choices, leading to sub-optimal outcomes. In contrast, LinUCB performs well when $\beta=1$ and $\beta=10$, demonstrating that its exploration-exploitation balancing strategy provides an advantage.

DYNUCB shows high regret in all scenarios. We believe that since it learns $\theta_i$ online, it obtains inaccurate $\theta_i$ in early rounds, leading to inaccurate clustering and consequently poor performance. In contrast, EWC algorithm utilizes the relatively accurate $\theta_i$ from offline training. Additionally, the loss-guided distance metric in clustering improves clustering performance.
\subsubsection{Ablation Study}
In this subsection, we perform an ablation study to assess the impact of each component of our proposed EWC algorithm. We aim to understand their contribution to the overall performance.
\tyeditnew{
EWC consists of three main components: (1) the user non-compliance model, (2) clustering and prediction with expert advice, and (3) linear regression on user context. 
\textit{Without non-compliance} is EWC algorithm without using the user non-compliance model. We use a linear model to learn user's preference parameter instead. 
\textit{Non-compliance} is EWC without clustering and prediction with expert advice. It is reduced to the user non-compliance model.
\textit{Without $u_i$} is EWC without using user context to accelerate preference learning.
We also incorporate oracle methods in this section to show the potential of our EWC algorithm.
\textit{Oracle Cluster} is EWC with precise clustering to integrate group behaviors into user decision-making. We use Oracle Cluster to test the learning speed of prediction with expert advice.
Lastly, \textit{Oracle $\boldsymbol{\theta}_{i}$} uses complete information of user-specific preferences learned by Algorithm \ref{Supervised_Preference_Learning} to test the user non-compliance model in the offline training of EWC.}

\begin{figure*}[!t]%
    \centering
    \hfill
    \subfloat[$\beta=0$]{\includegraphics[width=0.33\textwidth]{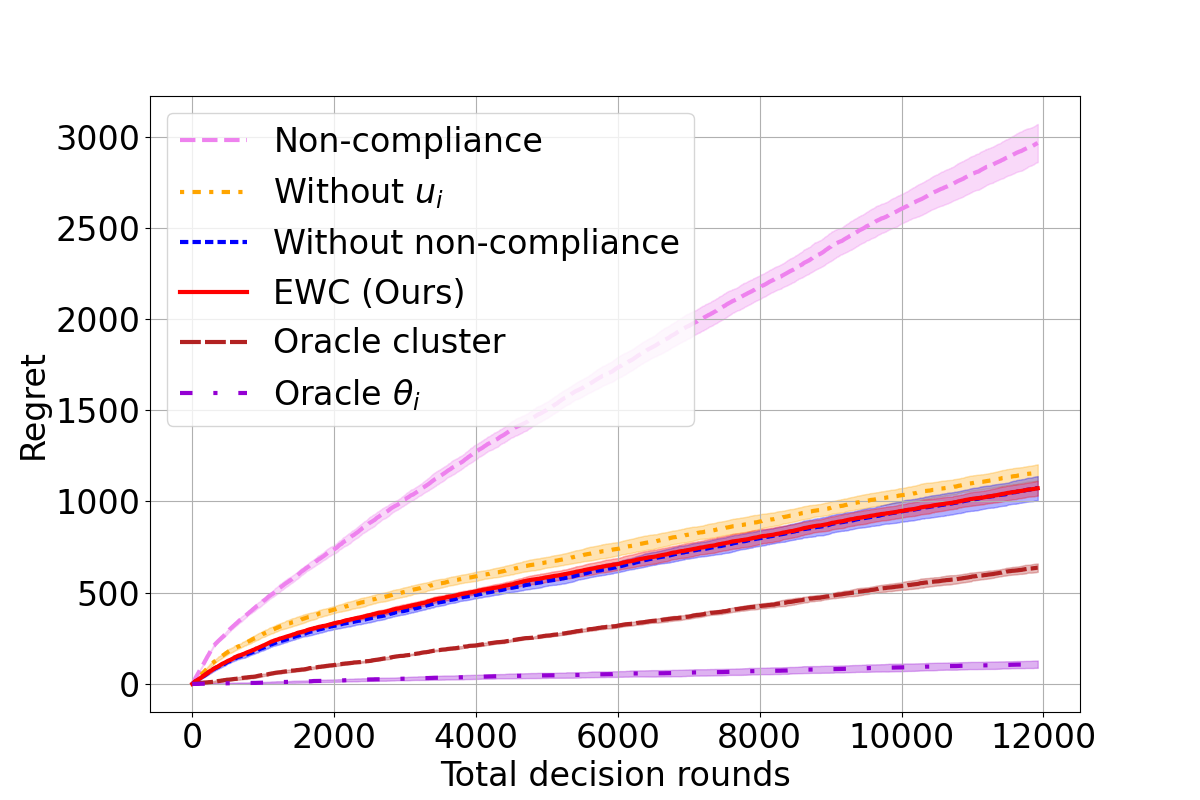}%
    \label{fig:beta=0_abla}}
    \hfill
    \subfloat[$\beta=1$]{\includegraphics[width=0.33\textwidth]{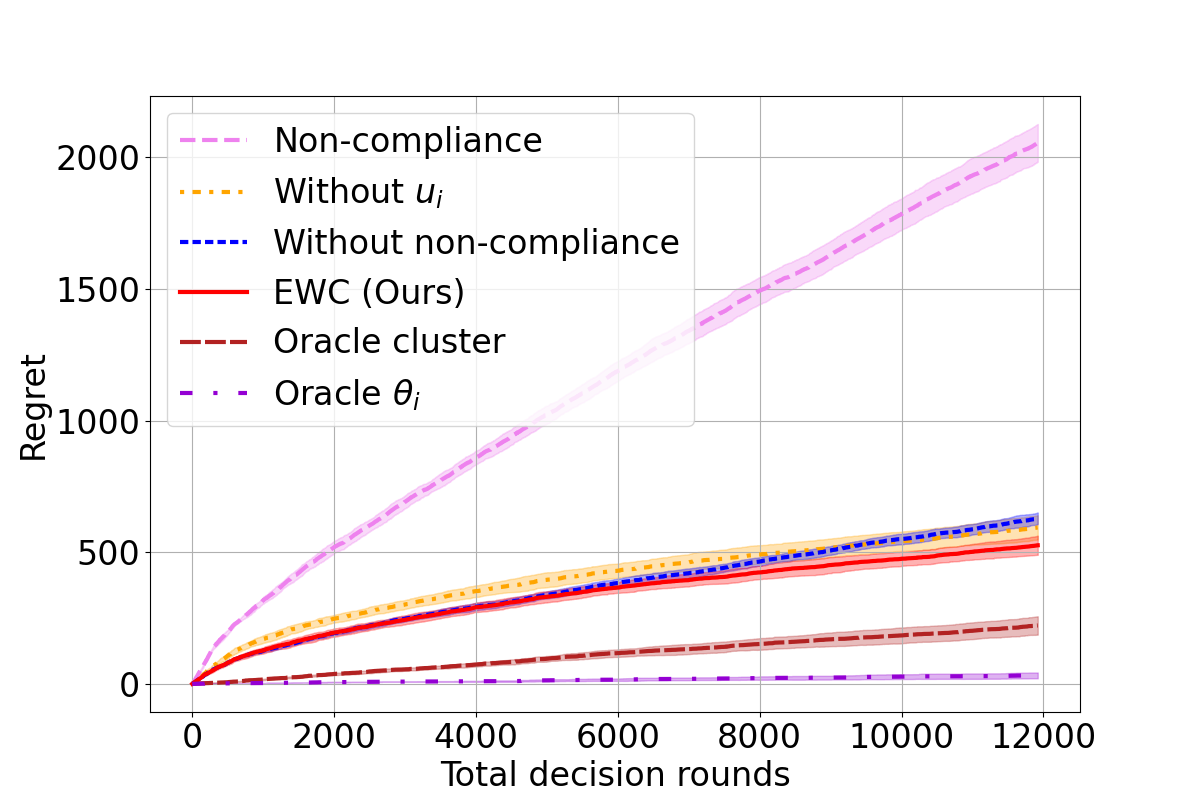}%
    \label{fig:beta=1_abla}}
    \hfill
    \subfloat[$\beta=10$]{\includegraphics[width=0.33\textwidth]{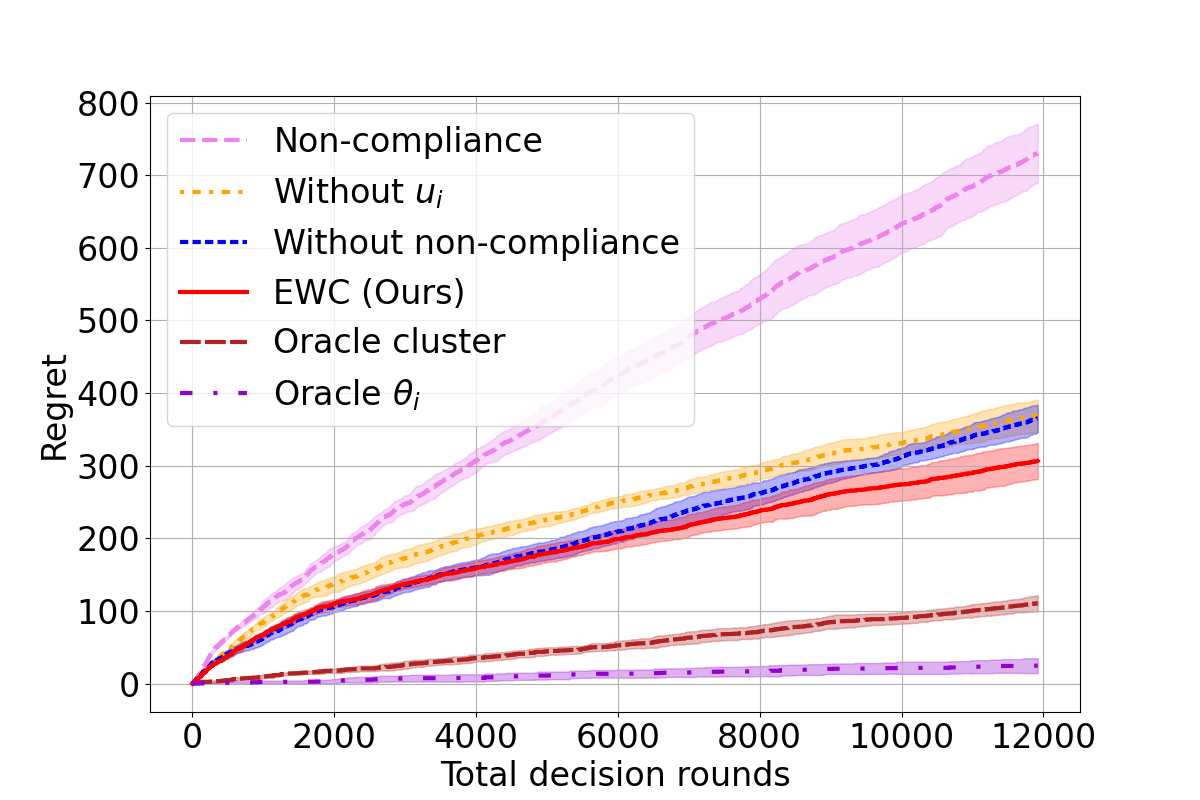}%
    \label{fig:beta=10_abla}}
    \hfill
    \caption{Ablation study of Expert with Clustering (EWC, Ours) in travel route recommendation. EWC consists of three main components: (1) the user non-compliance model, (2) clustering and prediction with expert advice, and (3) linear regression on user context. The approaches in the ablation study include Without non-compliance (EWC without (1)), Non-compliance (EWC without (2)), Without $u_i$ (EWC without (3)), Oracle cluster (EWC with oracle cluster centroids), and Oracle $\theta_i$ (EWC with oracle user preference parameters). The x-axis represents the total decision rounds, while the y-axis represents the regret. Lower regret indicates better performance. The comparison includes three scenarios with different values of $\beta$, where $\beta$ indicates the scale of users' dependence on recommendations in data generation. A higher $\beta$ means the population has a stronger preference for the recommended option. 
    }
    \label{fig:travel_abla}
\end{figure*}

\tyeditnew{Figure \ref{fig:travel_abla} shows the results of the ablation study. 
EWC achieves a much lower regret compared to Non-compliance. As explained in Section \ref{Results and Interpretation}, the clustering and prediction with expert advice components significantly accelerate preference learning. 
EWC exhibits a lower regret slope than Without $u_i$ in the early rounds, indicating that using user context leads to a good initialization of the weight of each expert, further decreasing regret.
EWC shows a lower slope than Without non-compliance across the entire time span, indicating that the user non-compliance model reduces the bias introduced by the anchoring effect when learning user preferences. 
The low slope of Oracle Cluster indicates that the shared preference in each cluster can represent each user's specific preference well. The long-term slope of EWC mirrors that of Oracle Cluster, suggesting that prediction with expert advice rapidly identifies each user's cluster identity.
The Oracle $\boldsymbol{\theta}_{i}$ shows extremely low regret due to its complete information of user-specific preference. It indicates that the user non-compliance model in the offline training successfully learns users' preferences.
These two methods show the potential lower regret bounds that EWC could aspire to achieve.
}

\vspace{-0.1in}
\subsection{Restaurant Recommendation}
\subsubsection{Data}
The dataset for restaurant recommendations were constructed using the Entree Chicago Recommendation Data \cite{misc_entree_chicago_recommendation_data_123}.
This rich dataset is a collection of user interactions with the Entree Chicago restaurant recommendation system, which includes user preferences, selections, and ratings of various dining establishments within the Chicago area. 
We select four features to be included in the option context: food quality, service level, price, and style. 
The detailed context description and parameter setting are shown in Appendix~\ref{details_exp}.

\subsubsection{Results}
\begin{figure}
    \centering
    \includegraphics[width=0.35\textwidth]{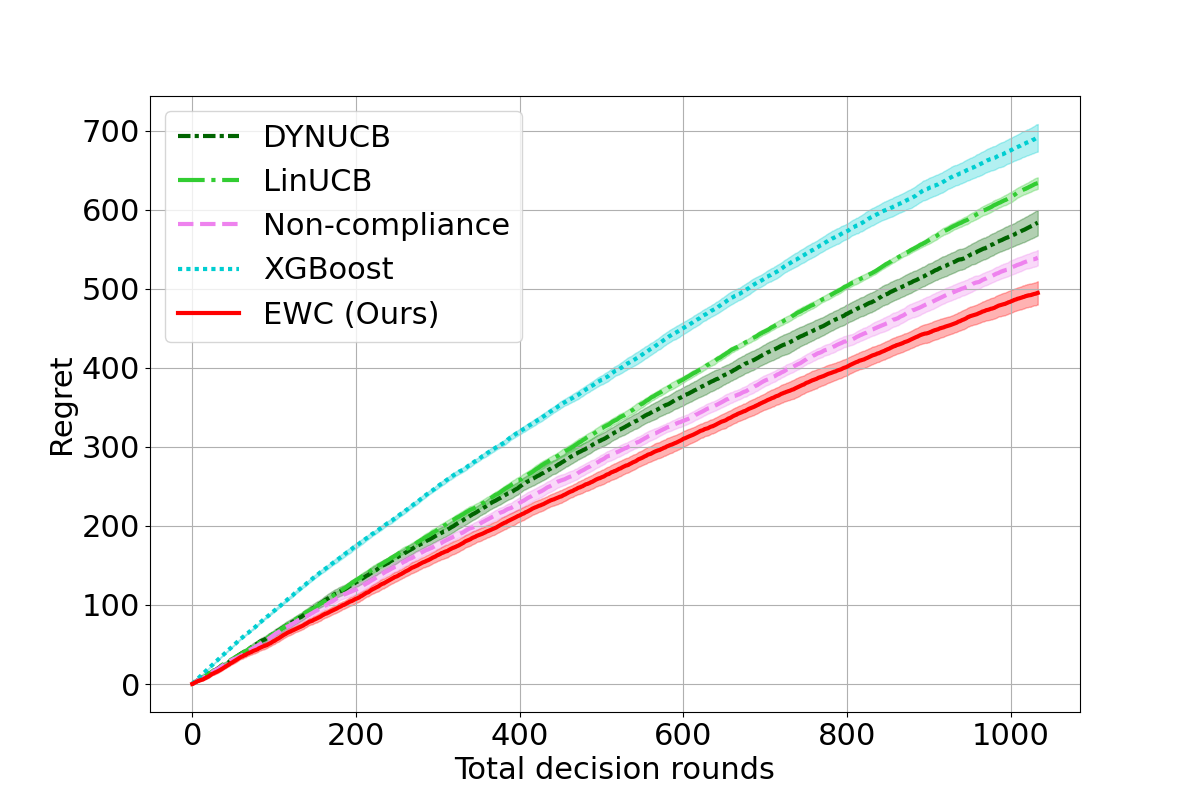}
    \caption{Regret of Expert with Clustering (EWC, Ours) and other baselines (XGBoost, LinUCB, DYNUCB, and the user non-compliance model) applied to restaurant recommendation data. The x-axis represents the total decision rounds, while the y-axis represents the regret. \tyeditnew{Lower regret indicates better performance.} EWC consistently shows lower regret than other baseline methods.}
    \label{fig:restaurant}
\end{figure}

The experiment is repeated with 10 different random seeds. Figure~\ref{fig:restaurant} provides the comparative result of EWC and baselines in the restaurant recommendation scenario over 1000 total rounds.
EWC algorithm shows the lowest regret across the entire time span, demonstrating its effectiveness in restaurant recommendation. XGBoost shows the highest regret, likely due to its cold start problem. XGBoost model is more complex than other baselines, but the limited number of decision rounds per user is insufficient for adequate training.
DYNUCB exhibits lower regret than LinUCB, and EWC outperforms the user non-compliance model, indicating that the clustering method efficiently leverages the hierarchical structure among users, thereby accelerating the preference learning process. The user non-compliance model performs better than LinUCB, demonstrating its adaptability to the Nah Bandit problem. This also contributes to EWC's superior performance compared to DYNUCB.

\vspace{-0.1in}
\section{Conclusion}
In this paper, we introduce a novel bandit framework—Nah Bandit. This framework offers the potential to accelerate preference learning.
To solve this problem efficiently, we first introduce a user non-compliance model that parameterizes the anchoring effect to reduce bias when learning user preferences. Based on the user non-compliance model, we introduce Expert with Clustering (EWC), a novel hierarchical contextual bandit algorithm designed to address the Nah Bandit. EWC efficiently utilizes both compliance and non-compliance from users, achieving low regret in different feedback scenarios. This work lays the foundation for future research in Nah Bandit.

However, EWC may not achieve sublinear regret in the long term because it uses the cluster centroid as an estimate of user-specific preferences each time.
\tyedit{In the future, we plan to address this by incorporating both individual user preferences and the shared preferences within each cluster. \tyyedit{Dynamically adjusting the number of clusters and refining cluster granularity may further enhance long-term accuracy.}}
\tyedit{Additionally, our EWC approach is designed to handle multi-dimensional data. However, clustering algorithms often face challenges in high-dimensional settings due to the "curse of dimensionality," which poses a bottleneck for our current method. To address this, we hope to incorporate the dimension reduction method into EWC.}
\tyedit{Moreover, the current user non-compliance model in EWC assumes a linear relationship between option context and user preferences. Extending this component with kernel methods or nonlinear models to better handle nonlinear relationships is a direction for future work.}
\vspace{-0.1in}
\section*{Acknowledgments}
This work was partially supported by the National Science Foundation (NSF) under grant number 2149548 and the Kwanjeong scholarship. The authors would like to thank Prof. Hamed Tabkhi and Babak Rahimi Ardabili for their survey data support, and Prof. Christos Cassandras and Prof. Andreas Malikopoulos for their insightful discussions.

{\appendices
\vspace{-0.1in}
\section{Proof details}
\subsection{Proof of Theorem \ref{sample_non-compliance}}\label{Proof_sample}
\begin{proof}
    Since we use the context of only two options to update the model where one is the user's choice and the other is not, and $\mathbf{p}_{i,t}=\boldsymbol{\sigma}(x_{i,t,1}^\intercal \boldsymbol{\theta}_i, x_{i,t,2}^\intercal \boldsymbol{\theta}_i)$, we can get $\mathbf{p}_{i,t}(1)=\sigma((x_{i,t,1}-x_{i,t,2})^\intercal\boldsymbol{\theta}_i)=\sigma(\tilde{x}_{i,t}^\intercal \boldsymbol{\theta}_i)$. Therefore, the user non-compliance model is equivalent to the logistic regression. 
    Since $\tilde{x}_{i,t}\sim\mathcal{N}(\mathbf{0},\mathbf{I}_d)$, $y_{i,t}|\tilde{x}_{i,t}\sim \text{Bern}(\sigma(\beta_i \tilde{x}_{i,t}^\intercal \theta_{i}^*))$, $\theta_i=\hat{\theta}_{\text{ERM}}(\{x_{i,t}, y_{i,t}\}_{t=1}^T)$, $\beta_i\geq \frac{4\sqrt{2\pi}}{\epsilon}$, and $T\geq \frac{C(d\log(1/\epsilon)+\log(1/\delta))}{\epsilon}$, by Lemma \ref{sample_LR}, with probability at least $1-\delta$, we have $||\theta_i-\theta^*_i||\leq \epsilon$.
\end{proof}
\vspace{-0.1in}
\subsection{Proof of Theorem \ref{Regret_EWC}}\label{Proof_EWC}
\tyedit{We first introduce the bound of clustering loss. The loss of K-Means is} $\mathcal{L}=\sum_{i=1}^N ||\mathbf{c}_{k(i)}-\boldsymbol{\theta}_{i}||^2 $, where $k(i)$ is the cluster centroid assigned to $\boldsymbol{\theta}_{i}$. 
\cite{K-MeansBound} proved the Uniform deviation bound of K-Means algorithm. 
Consider $P$ as any distribution on $\mathbbm{R}^d$ with mean $\boldsymbol{\mu} = \mathbbm{E}_P[\boldsymbol{\theta}_{i}]$ and variance $\sigma^2 =\mathbbm{E}_P[||\boldsymbol{\theta}_{i}-\boldsymbol{\mu}||^2]$. Assuming finite Kurtosis ($4^\text{th}$ moment) $\hat{M}_4<\infty$ and given $\epsilon \in (0,1)$, $\delta \in (0,1)$ and a sample size $m$ from $P$, we establish that for $ m\geq \frac{12800(8+\hat{M}_4)}{\epsilon^2\delta}\left(3+30K(d+4)\log 6K+\log \frac{1}{\delta}\right)$, \tyedit{with probability at least $1-\delta$:}
\begin{equation} 
    \label{K-MeansBound-Proof}
     | \mathcal{L}-\mathbbm{E}_P [\mathcal{L}]|\leq \frac{\epsilon}{2}\sigma^2 +\frac{\epsilon}{2}\mathbbm{E}_P [\mathcal{L}]
\end{equation}

\tyedit{Based on this, we prove the regret bound of EWC.}
\begin{proof}
\begin{equation}
\begin{aligned}
R_{\text{EWC}} =& \sum_{i=1}^N \sum_{t=1}^T \left( \langle \mathbf{p}_{i,t}, \mathbf{l}_{i,t}\rangle - 
    l(\hat{y}(\boldsymbol{\theta_i},\mathbf{x}_{i,t}), y_{i,t}) 
    \right)\\
                \end{aligned}
    \end{equation}
Since, $l(\hat{y}(\boldsymbol{\theta_i},\mathbf{x}_{i,t}), y_{i,t})=\mathbbm{1}_{\hat{y}(\boldsymbol{\theta_i},\mathbf{x}_{i,t})\neq y_{i,t}}=\frac{1}{2}||\mathbf{\hat{y}}(\boldsymbol{\theta_i},\mathbf{x}_{i,t})-\mathbf{y}_{i,t}||^2$,
    \begin{equation}
        \begin{aligned}
   R_{\text{EWC}} =& \sum_{i=1}^N \sum_{t=1}^T \left( \langle \mathbf{p}_{i,t}, \mathbf{l}_{i,t}\rangle - \frac{1}{2}||\mathbf{\hat{y}}(\boldsymbol{\theta_i},\mathbf{x}_{i,t})-\mathbf{y}_{i,t}||^2  \right)\\
    =& \sum_{i=1}^N \sum_{t=1}^T \left(\langle \mathbf{p}_{i,t}, \mathbf{l}_{i,t}\rangle - \frac{1}{2}||\mathbf{\hat{y}}(\mathbf{c}_{k^*(i)},\mathbf{x}_{i,t})-\mathbf{y}_{i,t}||^2  \right) \\
    + \frac{1}{2}\sum_{i=1}^N  \sum_{t=1}^T  &\left(||\mathbf{\hat{y}}(\mathbf{c}_{k^*(i)},\mathbf{x}_{i,t})-\mathbf{y}_{i,t}||^2-||\mathbf{\hat{y}}(\boldsymbol{\theta_i},\mathbf{x}_{i,t})-\mathbf{y}_{i,t}||^2  \right)\\
                \end{aligned}
    \end{equation}
Recall that $\frac{1}{2}||\mathbf{\hat{y}}(\mathbf{c}_k,\mathbf{x}_{i,t})-\mathbf{y}_{i,t}||^2 =l(\hat{y}(\mathbf{c}_k,\mathbf{x}_{i,t}),y_{i,t})=\mathbf{l}_{i,t}(k)$,
    \begin{equation}
        \begin{aligned}
    R_{\text{EWC}} =& \sum_{i=1}^N \sum_{t=1}^T \left(\langle \mathbf{p}_{i,t}, \mathbf{l}_{i,t}\rangle - \mathbf{l}_{i,t}(k^*(i)) \right) \\
    + &\frac{1}{2}\sum_{i=1}^N \left(||\mathbf{\hat{y}}(\mathbf{c}_{k^*(i)},\mathbf{X}_i)-\mathbf{y}_{i}||^2_F - ||\mathbf{\hat{y}}(\boldsymbol{\theta}_{i},\mathbf{X}_i)-\mathbf{y}_{i}||^2_F\right)\\
                \end{aligned}
    \end{equation}
    \tyedit{By the triangle inequality and the regret bound of Hedge algorithm \cite{HedgeRegretBound} that $\sum_{t=1}^T \left( \langle \mathbf{p}_{t},\mathbf{l}_{t} \rangle -\mathbf{l}_{t}(k^*) \right) \leq 2\sqrt{T \log K}$,}
    \begin{equation}
        \begin{aligned}
    R_{\text{EWC}}\leq & 2N\sqrt{T\log K}+\frac{1}{2}\sum_{i=1}^N||\mathbf{\hat{y}}(\mathbf{c}_{k^*(i)},\mathbf{X}_i)-\mathbf{\hat{y}}(\boldsymbol{\theta}_{i},\mathbf{X}_i)||^2_F\\
\end{aligned}
\end{equation}
By the Lipschitz condition, $\exists L$ s.t. $\forall i, \forall \boldsymbol{\theta}_1,\boldsymbol{\theta}_2$, $\frac{1}{T}||\mathbf{\hat{y}}(\boldsymbol{\theta}_1, \mathbf{X}_i)-\mathbf{\hat{y}}(\boldsymbol{\theta}_2,\mathbf{X}_i)||^2_F\leq L ||\boldsymbol{\theta}_1-\boldsymbol{\theta}_2||^2$
    \begin{equation}
        \begin{aligned}
        &\sum_{i=1}^N||\mathbf{\hat{y}}(\mathbf{c}_{k^*(i)},\mathbf{X}_i)-\mathbf{\hat{y}}(\boldsymbol{\theta}_{i},\mathbf{X}_i)||^2_F\\
            \leq &TL\sum_{i=1}^N ||\mathbf{c}_{k^*(i)}-\boldsymbol{\theta}_{i}||^2\\
            = &TL\mathcal{L}\leq TL (|\mathcal{L}-\mathbb{E}[\mathcal{L}]|+\mathbb{E}[\mathcal{L}])\\
            \end{aligned}
    \end{equation}
By inequation \ref{K-MeansBound-Proof}, with probability at least $1-\delta$, 
    \begin{equation}
        \begin{aligned}
            \tyedit{TL\mathcal{L}}\leq &TL\left(\frac{\epsilon}{2}\sigma^2+(\frac{\epsilon}{2}+1)\mathbb{E}[\mathcal{L}]\right)\\
        \end{aligned}
    \end{equation}
    \begin{equation}
        \begin{aligned}
        R_{\text{EWC}} \leq 2N\sqrt{T\log K}
        +\frac{1}{4}TL\left(\epsilon\sigma^2+(\epsilon+2)\mathbb{E}[\mathcal{L}]\right)
        \end{aligned}
    \end{equation}
\end{proof}
\vspace{-0.1in}
\subsection{Proof of Corollary \ref{EWC_Bound_GMM}}\label{Proof_EWC_Coro}
\begin{proof}
Since $\mathbf{c}_{k} = \boldsymbol{\mu}_k$, and $P=\sum_{k=1}^K \pi_k \mathcal{N}(\boldsymbol{\mu}_k, \Sigma_k)$, the expected squared distance $\mathbb{E}[||\boldsymbol{\theta}_{i}-\mathbf{c}_{k(i)}||^2]=\sum_{k=1}^K \pi_k \text{trace}(\Sigma_k)$. So,  $\mathbb{E}[\mathcal{L}]=N\mathbb{E}[||\boldsymbol{\theta}_{i}-\mathbf{c}_{k(i)}||^2]=N\sum_{k=1}^K \pi_k \text{trace}(\Sigma_k)$. Since $l_{\text{centroids}}=\frac{1}{4N}L\epsilon \sigma^2 +\frac{1}{4}L(\epsilon+2)\sum_{k=1}^K \pi_k \text{trace}(\Sigma_k)$, we can get $R_{\text{EWC}}\leq \overline{R}_{\text{EWC}}= 2N\sqrt{T\log K} +  TNl_{\text{centroids}}$.
\end{proof}
\vspace{-0.1in}
\subsection{Proof of Corollary \ref{AdvEWC}}\label{Proof_AdvEWC}
\begin{proof}
Since $\overline{R}_{\text{EWC}}=2N\sqrt{T\log K}+TNl_{\text{centroids}}$ and $\overline{R}_{\text{LinUCB}}= CN\sqrt{Td\log^3{(AT\log T/\delta})}$, $\overline{R}_{\text{EWC}}<\overline{R}_{\text{LinUCB}}$ is equivalent to $\sqrt{T}l_{\text{centroids}}<C\sqrt{d\log^3{(AT\ln T/\delta})}-2\sqrt{\log K}$. Since $K\ll T$, when $\sqrt{T}l_{\text{centroids}}<C-2$, we can get $\overline{R}_{\text{EWC}}<\overline{R}_{\text{LinUCB}}$.
\end{proof}
\vspace{-0.1in}
\section{Experimental Setup Details}\label{details_exp}
In the travel route recommendation, the contexts include travel time and CO2 emission of different routes: the regular route and the eco-friendly route. The context of the regular route is $[100, 100]$, which means $100\%$ of travel time and CO2 emission. The eco-friendly route has higher travel time and lower CO2 emission compared to the regular one.
We generate the training and testing data based on the survey data. The user preference parameter $\theta_i$ is initially sampled from a multivariate normal distribution with mean $[-0.1, -0.1]$ and covariance $\text{Diag([0.01, 0.01])}$. We assume $\theta_{\text{rec}}$ also shows cluster characteristic, so we sample $\theta_{\text{rec}}$ from a beta distribution $B(0.3, 0.3)$. Each dimension of the option context for the eco-friendly route is generated from a normal distribution, with mean $[104.29, 91.99]$ and standard deviation $[5.62, 4.06]$. \tyedit{The parameters of travel route recommendation are listed in Table \ref{tab:para_travel_rec}, and the description of context is shown in Table \ref{tab:context_travel_rec}.}
The user context ``Age" and ``Education level" are transformed into one-hot encodings, while others are binary variables.

The parameters used in the restaurant recommendation experiment are listed in Table \tyedit{\ref{tab:para_travel_rec}}. The description of the option context is shown in Table \tyedit{\ref{tab:context_travel_rec}}. 
The user context 'Style' is transformed into one-hot encoding. `Food quality' and `Service level' are transformed to $(0, 0.25, 0.5, 0.75, 1)$. 'Price' is transformed to $(0, 0.33, 0.67, 1)$ accordingly.
\vspace{-0.05in}
\begin{table}[ht]
\centering
\caption{\tyedit{A list of parameters and their values in travel route recommendation and restaurant recommendation.}} 
\label{tab:para_travel_rec}
\begin{tabular}{|l|c|c|}
\hline
\textbf{Parameter} & \textbf{Travel} & \textbf{Restaurant} \\
\hline
\textbf{General Parameters} & & \\
Number of decision rounds \( T \) & 40 & 3--105 \\
User context dimensions \( D \) & 9 & -- \\
Option context dimensions \( d \) & 2 & 9 \\
Number of options \( A \) & 2 & 2--18 \\
\hline
\textbf{Training Parameters} & & \\
Number of users \( N^\prime \) & 446 & 188 \\
Learning rate & 0.5 & 0.5 \\
\(L_2\) regularization parameter & 0.001 & 0.01 \\
Number of clusters \( K \) & 6 & 8 \\
\hline
\textbf{Testing Parameters} & & \\
Number of users \( N \) & 298 & 75 \\
Exploration rate for LinUCB & 0.05 & 0.05 \\
Learning rate for EWC & 1 & 1 \\
\hline
\end{tabular}
\end{table}
\vspace{-0.1in}
\begin{table}[ht]
\centering
\caption{\tyedit{Description of user and option context in travel route recommendation and restaurant recommendation.}} 
\label{tab:context_travel_rec}
\begin{tabular}{|l|c|c|}
\hline
\textbf{Travel}& \tyeditnew{\textbf{Range}} &\textbf{Dimensions} \\
\hline
\textbf{User context} & & \\
Age  & \{18-34, 35-49, 50-64\}& 3\\
Gender& \{Male, female\}& 1\\
Education level& \makecell{\{High school, bachelor,\\ master or higher\}} & 3\\
Number of cars & \{One, two or more\}&1\\
\hline
\textbf{Option context} & & \\
Travel time & \tyedit{$\mathbb{R}_{\geq0}$} &1\\
CO2 emission & \tyedit{$\mathbb{R}_{\geq0}$} &1\\
\hline
\textbf{Restaurant} & \tyeditnew{\textbf{Range}}& \textbf{Dimensions} \\
\hline
\textbf{Option context} & & \\
Food quality& \makecell{\{Fair, good, excellent,\\ extraordinary, near-perfect\}}& 1\\
Service level&\makecell{\{Fair, good, excellent,\\ extraordinary, near-perfect\}}& 1\\
Price& \makecell{\{Below \$15, \$15-\$30,\\ \$30-\$50, over \$50\}}& 1\\
Style&\makecell{\{American, Asian, Latin,\\ Middle Eastern, other\}}& 6\\
\hline
\end{tabular}
\end{table}
\bibliographystyle{ieeetr}
\bibliography{reference}
\vspace{-0.05in}
\begin{IEEEbiography}[{\includegraphics[width=1in,height=1.25in,clip,keepaspectratio]{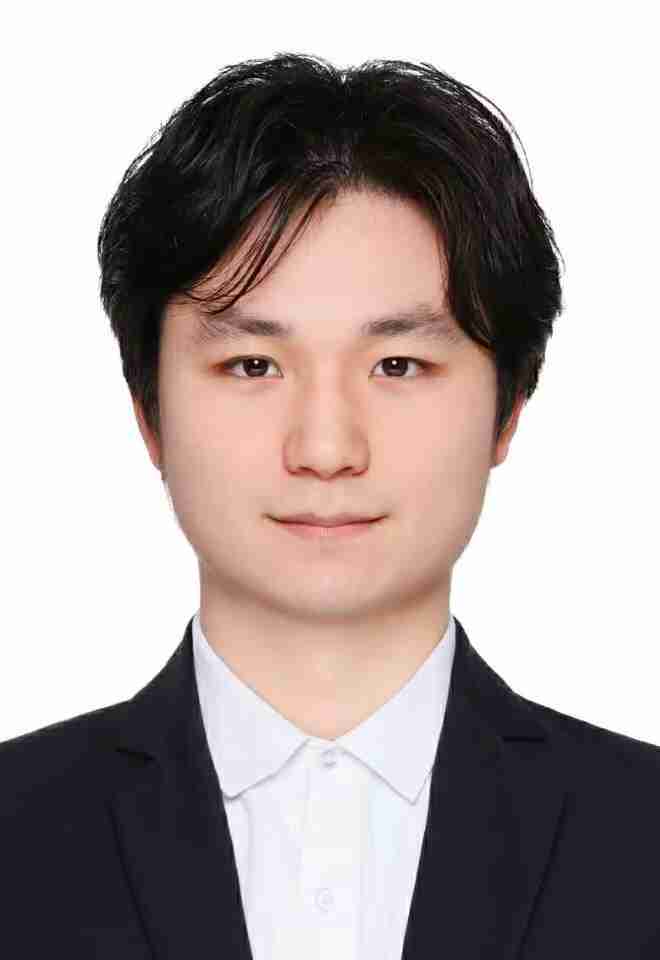}}]{Tianyue Zhou} is a Ph.D. candidate in Civil and Environmental Engineering at the Massachusetts Institute of Technology (MIT). He earned his B.S. degree in Computer Science from ShanghaiTech University. His research interest is machine learning. He aims to develop sample-efficient machine learning algorithms for solving practical control problems. 
\end{IEEEbiography}
\vspace{-33pt}
\begin{IEEEbiography}[{\includegraphics[width=1in,height=1.25in,clip,keepaspectratio]{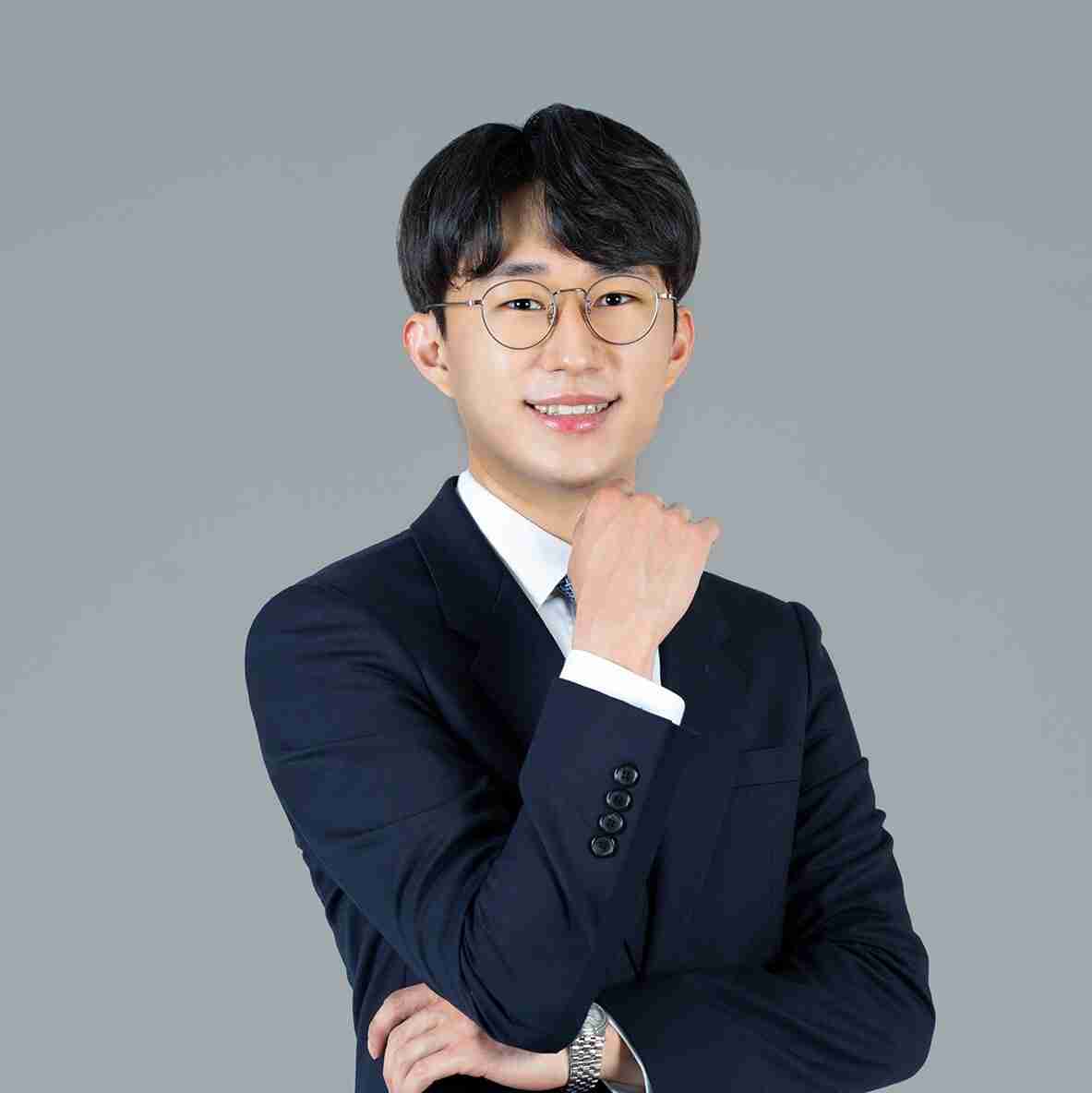}}]{Jung-Hoon Cho}
is a Ph.D. candidate in Civil and Environmental Engineering at the Massachusetts Institute of Technology (MIT). He earned both his M.S. and B.S. degrees in Civil and Environmental Engineering from Seoul National University. His primary research interest lies at the intersection of transportation and machine learning. Jung-Hoon aims to develop generalizable machine learning models to optimize traffic flow, thereby reducing urban congestion and greenhouse gas emissions.
\end{IEEEbiography}
\vspace{-33pt}
\begin{IEEEbiography}[{\includegraphics[width=1in,height=1.25in,clip,keepaspectratio]{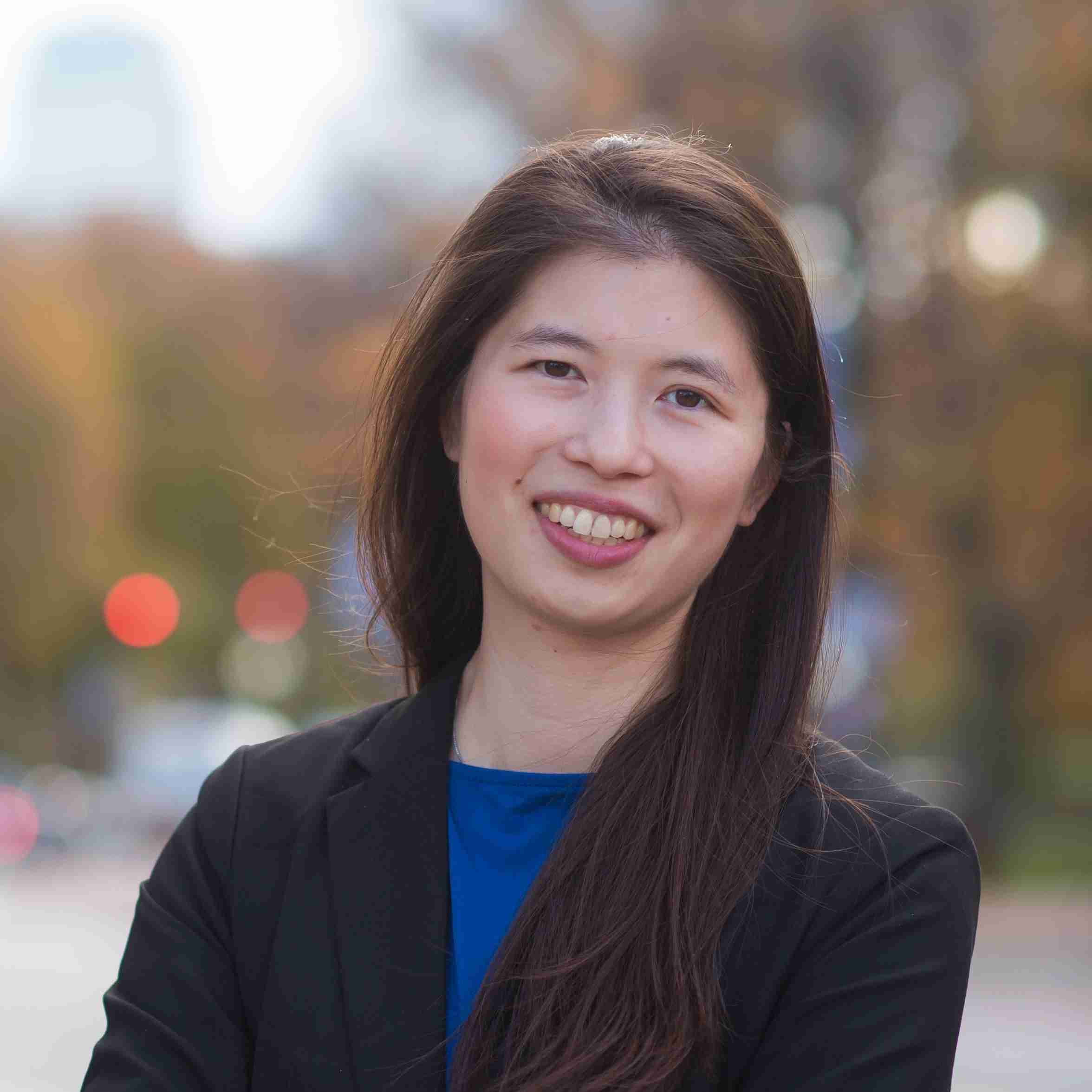}}]{Cathy Wu}
is an Associate Professor at MIT in LIDS, CEE, and IDSS. She holds a Ph.D. from UC Berkeley, and B.S. and M.Eng. from MIT, all in EECS, and completed a Postdoc at Microsoft Research. Her research interests are at the intersection of machine learning, autonomy, and mobility. Her research aims to advance generalizable optimization to enable next-generation mobility systems. Cathy is the recipient of the NSF CAREER, several PhD dissertation awards, and several publications with distinction. She serves on the Board of Governors for the IEEE ITSS and as an Area Chair for ICML and NeurIPS.
\end{IEEEbiography}

\vspace{11pt}
\vfill
\end{document}